\documentclass{article}

\PassOptionsToPackage{numbers, sort&compress}{natbib}
\usepackage[preprint,nonatbib]{neurips_2025}

\usepackage[utf8]{inputenc} % allow utf-8 input
\usepackage[T1]{fontenc}    % use 8-bit T1 fonts
\usepackage{hyperref}       % hyperlinks
\usepackage{url}            % simple URL typesetting
\usepackage{booktabs}       % professional-quality tables
\usepackage{amsfonts}       % blackboard math symbols
\usepackage{nicefrac}       % compact symbols for 1/2, etc.
\usepackage{microtype}      % microtypography
\usepackage[capitalize]{cleveref}
\crefname{section}{Sec.}{Secs.}
\Crefname{section}{Section}{Sections}
\Crefname{table}{Table}{Tables}
\crefname{table}{Tab.}{Tabs.}
\usepackage{graphicx} % 用来显示图片和表格
\usepackage{wrapfig}
\usepackage{caption} % 表6标题正常显示
\usepackage{multirow}
\usepackage{float} % 这个包和suppl里的标题首图直接相关
\usepackage[style=ieee, backend=biber]{biblatex}
\usepackage[table,xcdraw]{xcolor}
\usepackage{array}
\usepackage[shortlabels,inline]{enumitem}
\title{Efficient and Comprehensive Feature Extraction in Large Vision-Language Model for Pathology Analysis}

\author{
Shengxuming Zhang$^{1}$ \quad Weihan Li$^{1}$ \quad Tianhong Gao$^{1}$ \quad Jiacong Hu$^{2}$ \quad Haoming Luo$^{1}$ \\ \textbf{Xiuming Zhang}$^{3}$\thanks{Corresponding authors, email: \texttt{\{xm\_zhang, jzhang1989, brooksong, zunleifeng\}@zju.edu.cn}} \quad
\textbf{Jing Zhang}$^{3}$\footnotemark[1] \quad
\textbf{Mingli Song}$^{2}$\footnotemark[1] \quad
\textbf{Zunlei Feng}$^{1,2}\footnotemark[1]$ \\
$^{1}$School of Software Technology, Zhejiang University\\
$^{2}$College of Computer Science and Technology, Zhejiang University\\
$^{3}$First Affiliated Hospital, College of Medicine, Zhejiang University
}

\begin{document}

\maketitle

\begin{abstract}
Pathological diagnosis is vital for determining disease characteristics, guiding treatment, and assessing prognosis, relying heavily on detailed, multi-scale analysis of high-resolution whole slide images (WSI).
However, existing large vision-language models (LVLMs) are limited by input resolution constraints, hindering their efficiency and accuracy in pathology image analysis.
To overcome these issues, we propose two innovative strategies: the mixed task-guided feature enhancement, which directs feature extraction toward lesion-related details across scales, and the prompt-guided detail feature completion, which integrates coarse- and fine-grained features from WSI based on specific prompts without compromising inference speed.
Leveraging a comprehensive dataset of 490K samples from diverse pathology tasks, we trained the pathology-specialized LVLM, OmniPath. Extensive experiments demonstrate that this model significantly outperforms existing methods in diagnostic accuracy and efficiency, providing an interactive, clinically aligned approach for auxiliary diagnosis in a wide range of pathology applications.
\end{abstract}

% ################# Begin of main text #################

% ############ Introduction ############
\section{Introduction}
\label{sec:intro}

\begin{figure*}[t]
    \centering
    \includegraphics[width=0.85\textwidth]{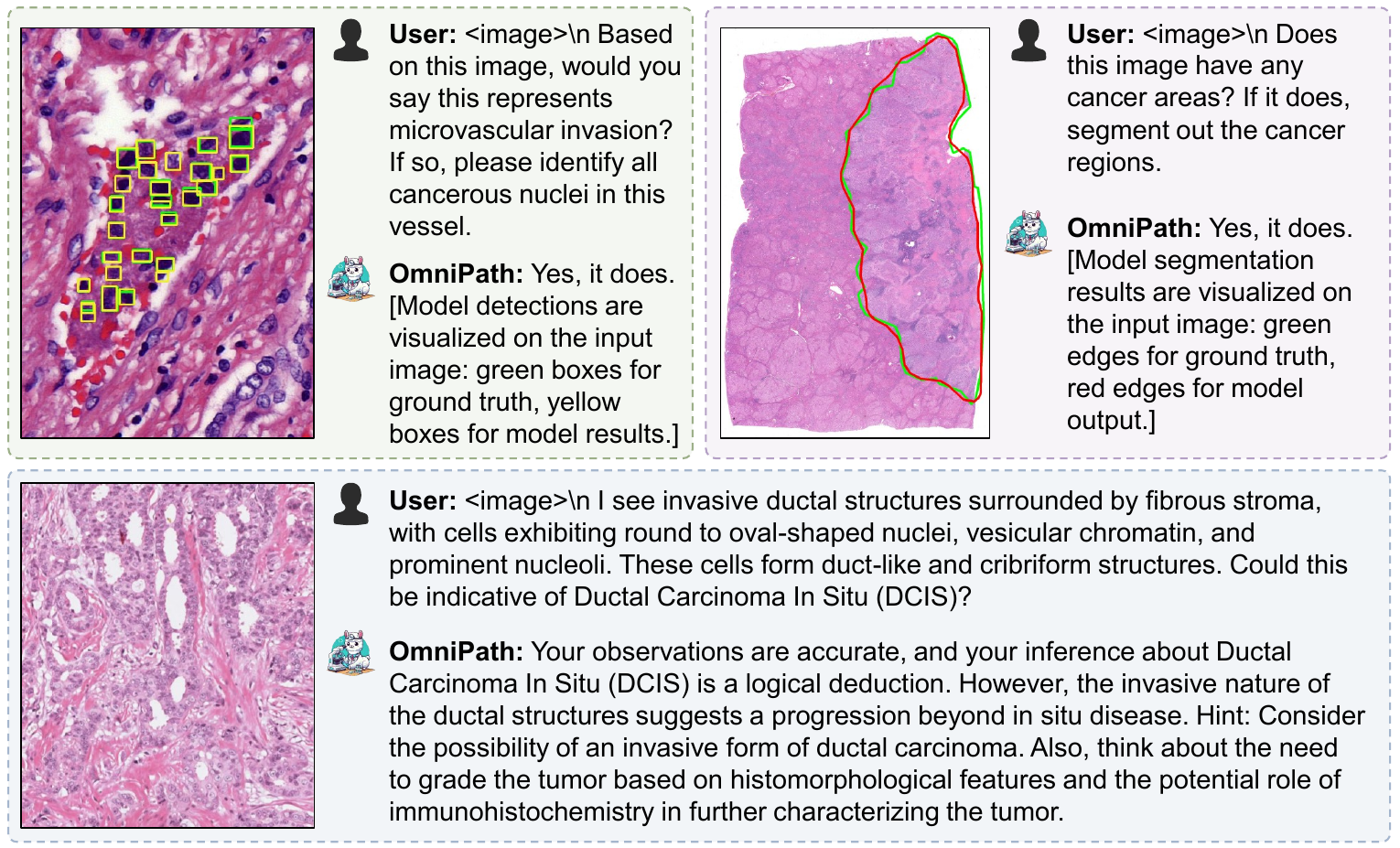}
    \vspace{-0.5em}
    \caption{
    Dialogue examples of our OmniPath, a vision-language model optimized for pathology, applied to referring expression detection, segmentation, and visual question answering. Notably, in the first example, OmniPath is tasked with detecting cancer cell nuclei within blood vessels. Results show that OmniPath accurately identifies most nuclei within vessels without mistakenly detecting any outside, demonstrating its capability to understand pathological concepts and reason effectively.
    }
    \label{fig:sample}
    \vspace{-1.7em}
\end{figure*}

Pathological diagnosis, as the ``gold standard'' of disease diagnosis, holds an irreplaceable central position in clinical diagnostics. Through microscopic morphological examination of patient tissues and cells, it not only determines the nature, type, and staging of diseases but also provides critical information for clinical treatment planning, prognosis assessment, and efficacy monitoring. The emergence and development of digital pathology are transforming this traditional field. By using high-resolution scanning equipment to convert glass slides into whole slide images (WSI), digital pathology overcomes the limitations of conventional pathology that rely on microscopy, enabling remote consultations and real-time consultations while paving new paths for medical education, research collaborations, and long-term clinical data storage.

WSI is characterized by ultra-high resolution, with single images typically exceeding 50,000 × 50,000 pixels. This high resolution allows WSI to encompass a wealth of multi-scale features, from organ-level structures to cellular-level details. However, only a subset of these features is directly relevant to disease diagnosis. Pathologists must therefore observe the slides at multiple magnifications to comprehensively capture morphological characteristics of lesions at various scales, identifying the key features closely related to diagnosis to avoid misdiagnosis or missed diagnoses.

In the field of digital pathology, artificial intelligence is playing an increasingly vital role~\cite{feng2021edge,feng2021mutual,Zhang_2023_CVPR}. Many pure vision deep learning models have been developed to assist in the diagnosis of pathology WSIs~\cite{chen2022scaling,UNI,xu2024whole,vorontsov2024foundation,wang2024pathology}. With the rapid development of large language models (LLM)~\cite{gpt4,llama3,qwen25,deepseekr1} and large vision-language models (LVLM)~\cite{llava15,qwen2vl,shikra,lisa}, these models have shown substantial auxiliary capabilities across various domains, and recent research~\cite{pathasst,pathmmu,quiltllava,pathchat,pathalign,pallava,pathgen,prism} has attempted to apply LVLM to pathology.

However, existing methods still face significant limitations: traditional pure vision models require WSIs to be divided into thousands of patches, with an encoder network extracting features from each patch and an aggregator network synthesizing the final result. These methods inevitably extract numerous redundant features, leading to a prolonged diagnostic process. Current pathological LVLMs, due to input constraints, can only process either single pathological image patches or low-resolution thumbnails of WSI. While this improves processing speed, it either lacks global information for single patches or loses substantial detail information for WSI, making it difficult to meet clinical assisted diagnostic requirements.
Furthermore, experimental analysis (given in \cref{sec:analysis}) of image tokens that LVLM focuses on for decision-making reveals that existing LVLMs often overly emphasize the features of a few key tokens in the input image.
While this feature extraction pattern can summarize image content, it fails to comprehensively capture multi-scale features related to lesions, thus affecting the model's diagnostic performance.

To enhance the accuracy and reliability of intelligent pathological diagnosis and analysis, we develop an efficient and comprehensive feature extraction scheme specifically tailored for LVLM in pathology, providing complete, multi-scale feature support for various types of pathological diagnostic analysis tasks.
In addressing the issue of the model focusing only on a few key tokens of the image, we introduce the mixed task-guided feature enhancement (MGFE) strategy. Through adding instruction-following data for detecting and segmenting diverse pathological concepts, coupled with corresponding model module improvements, we enhance the model's ability to perceive lesion-related detailed features across the whole image while achieving full coverage of visual task types in pathological analysis.
Furthermore, given the need for multi-scale features in pathology slide analysis, we design the prompt-guided detail feature completion (PGFC) strategy. This strategy first captures the coarse-grained global features of a WSI and then, based on specific task requirements provided by prompts, extracts fine-grained features from key focus areas. By merging coarse- and fine-grained features, this approach enhances accuracy across tasks while avoiding the input of exhaustive detail features, thereby maintaining high inference speed.

To make LVLM truly applicable to clinical pathology for auxiliary diagnosis, we curate visual instruction-following data for multiple tasks from several institutions, based on diagnostic items in actual pathology reports.
These tasks include cancer region detection and segmentation, cancer grading and subtyping, identification of vascular and neural invasion, and lymph node metastasis detection, among others.
Furthermore, to strengthen the model’s understanding of foundational pathology concepts, we collected training data for fundamental tasks such as nucleus detection and classification, vascular and neural detection, lymph node detection, and tumor-infiltrating lymphocyte identification.
Additionally, we integrated pathology image-text data from publicly available online resources, including the PubMed database~\cite{arch,pathasst}, pathology textbooks and atlases~\cite{pathvqa,pathasst}, The Cancer Genome Atlas (TCGA)~\cite{pathgen,wsivqa}, Twitter posts~\cite{openpath,pathmmu}, and educational histopathology videos on YouTube~\cite{quilt1m,quiltllava}.
This integration yielded a comprehensive dataset covering 21 organs with approximately 490K training samples. Leveraging our efficient and comprehensive feature extraction scheme and this extensive dataset, we trained OmniPath, a specialized LVLM for pathology, capable of providing comprehensive pathology auxiliary diagnostic services through human-computer interaction. Extensive experimental results demonstrate that OmniPath outperforms existing pathology LVLMs across a range of diagnostic tasks, better aligning with the actual needs of clinical practice.

In conclusion, the main contributions of our work are summarized as follows:
\begin{itemize}[left=0pt,labelindent=\parindent]

\item The mixed task-guided feature enhancement strategy is devised to direct large vision-language models to capture detailed pathology image features through fine-grained tasks targeting local features, while incorporating model module improvements. This approach reduces the model's overreliance on global features represented by a few key image tokens.

\item The prompt-guided detail feature completion strategy is devised to supplement key region detail features based on specific task requirements, significantly improving the accuracy of various pathology slide analysis tasks while maintaining high inference efficiency.

\item We developed OmniPath, a pathology-specific large vision-language model trained on a multi-source dataset encompassing 21 organs with 490K samples. Extensive experiments demonstrate OmniPath's superior performance over existing models across diagnostic tasks, aligning with clinical needs through an interactive framework.

\end{itemize}

% ############ Related Work ############
\vspace{-1em}
\section{Related Work}
\label{sec:related}

\vspace{-0.3em}
\paragraph{Pure vision deep models} have long been used in pathology image analysis, initially focusing on specialized architectures for tasks like nuclei segmentation~\cite{feng2021mutual}, vessel segmentation~\cite{feng2021edge}, and microvascular invasion detection~\cite{Zhang_2023_CVPR}. Broader WSI analysis tasks, such as cancer subtyping and prognosis prediction, often employ multiple instance learning (MIL)~\cite{abmil,transmil2021shao,camel2019xu,thandiackal2022differentiable}, which partitions WSIs into numerous patches for feature extraction and aggregation. To enhance generalization, recent work leverages self-supervised training on large-scale unlabeled WSIs~\cite{chen2022scaling,UNI,xu2024whole,vorontsov2024foundation,wang2024pathology}, improving overall performance and rare disease identification. However, this approach remains computationally intensive and risks diluting critical pathological features with irrelevant information.

\vspace{-0.7em}
\paragraph{Large vision-language models} have recently been explored for pathology image analysis~\cite{pathasst,pathmmu,quiltllava,pathchat,pathalign,pallava,pathgen,prism,pathinsight}, typically using LLaVA-based architectures~\cite{llava} fine-tuned with instruction-following data from diverse sources. PathAsst~\cite{pathasst} trained a CLIP~\cite{clip} model with PubMed and internal image-caption pairs, using ChatGPT~\cite{instructgpt} to generate more complex instructions. Quilt-LLaVA~\cite{quiltllava} extracted pathology concepts from YouTube tutorials via mouse pointer trails and constructed training data with GPT-4~\cite{gpt4}. PathMMU~\cite{pathmmu} compiled multi-source data for pathology visual QA, with expert-reviewed test sets. PathAlign~\cite{pathalign} used a BLIP-2~\cite{blip} Q-Former to extract WSI features for LLMs, while PA-LLaVA~\cite{pallava} introduced a scale-invariant connector to mitigate image resizing losses.

However, most models are limited to standard-sized images, processing only single patches or low-resolution WSI thumbnails, leading to a loss of global context or fine details. While PathAlign~\cite{pathalign} supports full WSI input, it still requires per-patch feature extraction. Additionally, these models are mainly suited for image description and visual QA but lack capabilities for fine-grained tasks like detection and segmentation, as well as complex multi-step diagnostic reasoning.

% ############ Analysis ############
\section{Analysis of Drawback in Existing LVLMs}
\label{sec:analysis}

\subsection{Preliminaries}
Today's most prominent open-source LVLM, like LLaVA~\cite{llava}, successfully integrate vision and language capabilities. For input pair $x = (x_v, x_t)$, where $x_v$ represents the image and $x_t$ represents the text prompt, the model first processes them through two embedding networks: vision encoder $\mathcal{V}$, consisting of a CLIP-based  Vision Transformer (ViT)~\cite{clip} followed by a projection layer, maps the image into feature embedding $e_v = (e_v^1, ..., e_v^N)$ where $N$ is the number of image tiles, while text encoder $\mathcal{T}$, comprising a tokenizer and an embedding layer, transforms the text into feature embedding $e_t = (e_t^1, ..., e_t^M)$ where $M$ is the number of text tokens. Both embeddings lie in the same feature space and serve as input to the large language model $\mathcal{M}$, which generates the output sequence $y$. The model can be formalized as: $y = \mathcal{M}(e_v, e_t)$, where $e_v = \mathcal{V}(x_v)$ and $e_t = \mathcal{T}(x_t)$.

\begin{wrapfigure}{r}{0.5\textwidth}
    \centering
    \vspace{-1.4em}
    \includegraphics[width=0.5\textwidth]{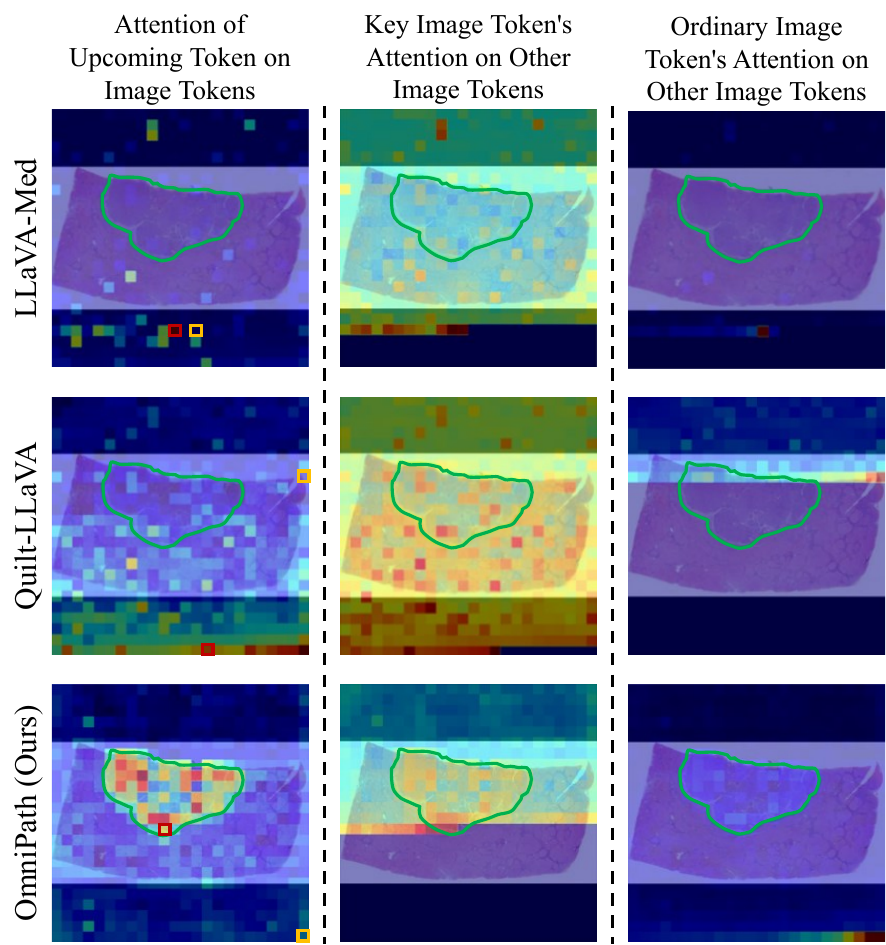}
    \vspace{-1.2em}
    \caption{The green contours on the pathology slides mark cancerous regions annotated by pathologists. The first column shows the attention distribution heatmap of the LLM's final input token over all image tokens, where the intensity of attention values is mapped from blue (low) to red (high). In each row showing different model results, a red box and a yellow box are used to select a key token (with relatively high attention) and an ordinary token (with relatively low attention) respectively. The attention distributions of the selected key token and ordinary token over other image tokens are then visualized in the second and third columns respectively. All experiments were conducted using identical prompts, with attention values extracted from the first layer of the LLM.}
    \label{fig:attn}
    \vspace{-0.5em}
\end{wrapfigure}

\subsection{Decision-Dependent Image Tokens Analysis}
\label{sec:heatmap_analysis}
To further investigate the image feature patterns that LVLM relies on during the answer generation and decision-making processes, and to optimize the model to focus more effectively on task-relevant features, thereby improving accuracy in responding to human queries, we conducted an attention pattern analysis on existing medical LVLMs. Specifically, we selected two representative models, LLaVA-Med~\cite{llavamed} and Quilt-LLaVA~\cite{quiltllava}, using a unified prompt, \textit{``What cancer subtype is shown in this image?''} to guide the models in performing cancer subtype identification on pathology slides. Our proposed OmniPath model was included as a comparison. Visualization of the relevant analysis results is shown in \cref{fig:attn}.

We extracted the attention matrix from the input layer of $\mathcal{M}$ and averaged the attention values across all heads to obtain matrix $\Psi \in \mathbb{R}^{(N+M)\times(N+M)}$. In this matrix, the $i$-th row of $\Psi$ represents the attention distribution of the $i$-th embedding token in the input of $\mathcal{M}$ towards other tokens. Since a decoder-only Transformer model is used, $\Psi$ takes the form of a lower triangular matrix. To analyze the relationship between the upcoming generated content and the image tokens, we selected the attention values of the final embedding token towards all image tokens, denoted as \[\Psi_{(N+M), e_v} \in \mathbb{R}^{N},\] and restored it to a two-dimensional representation with the same shape as the original image feature. We then generated a heatmap and overlaid it on the input image for visualization, as shown in the first column of \cref{fig:attn}.

In the heatmap, attention intensity is mapped from blue (low) to red (high). Cancerous regions in the input pathology slide $x_v$ are annotated by pathologists with green contours. Ideally, to achieve accurate cancer subtype identification, the model should focus on image features within the cancerous region rather than on other tissue and background areas. However, the heatmaps for LLaVA-Med~\cite{llavamed} and Quilt-LLaVA~\cite{quiltllava} reveal that only a few key image tokens receive high attention weights, and these key tokens are primarily located outside the cancerous regions.

To further analyze the differences in the image information encoded by these key image tokens and other ordinary image tokens, we selected one key image token \( e_v^{k} \) and one ordinary token \( e_v^{o} \) from each experiment, and visualized their attention distributions over all image tokens, denoted as \( \Psi_{e_v^{k}, e_v} \in \mathbb{R}^{N} \) and \( \Psi_{e_v^{o}, e_v} \in \mathbb{R}^{N} \), respectively. The selected key and ordinary tokens are indicated in the first column of \cref{fig:attn} with red and yellow boxes, respectively. The corresponding attention heatmap visualizations are presented in the second and third columns of \cref{fig:attn}.

Through visual analysis, it can be observed that the key image token $e_v^{k}$ exhibits high attention values across all preceding image tokens, whereas the ordinary token $e_v^{o}$ focuses only on nearby preceding tokens. This indicates that the primary function of the key token is to aggregate and distill the global semantic information of the entire image for use by the LLM~$\mathcal{M}$. However, this mechanism has limitations: $\mathcal{M}$ can only obtain a coarse-grained conceptual representation of the image, which may not only include redundant background information but, more critically, miss essential local lesion features and spatial structure information. This directly results in suboptimal performance of existing pathology LVLMs on diagnostic analysis tasks for pathology WSIs.

In contrast, in the optimized OmniPath model, the heatmap of $\Psi_{(N+M), e_v}$ shows that key image tokens are concentrated in the cancerous regions, indicating that these areas contribute more information to $\mathcal{M}$, aiding in accurate cancer subtype identification. From the heatmap of $\Psi_{e_v^{k}, e_v}$, it can be seen that although the key token maintains high attention values across all preceding image tokens, its focus on tokens within the cancerous regions is significantly higher. This demonstrates that OmniPath can more precisely capture critical diagnostic features in pathology WSIs, thus achieving a better response to user instructions. The next section will elaborate on the optimization strategies employed in OmniPath.

% ############ Method ############

\section{Method}
\label{sec:method}

To address the identified limitations of existing LVLMs, specifically their tendency to over-rely on key tokens and inability to comprehensively capture multi-scale pathological features, we propose a novel framework that enhances both feature extraction precision and efficiency. Our approach consists of two complementary strategies: the mixed task-guided feature enhancement (MGFE) and the prompt-guided detail feature completion (PGFC). These strategies work in concert to improve the model's capability in pathological image analysis by targeting the core challenges revealed in our previous analysis while maintaining computational efficiency and diagnostic accuracy.
OmniPath trained with these two strategies is shown in \cref{fig:framework}.
Below, we elaborate on these strategies and their implementation.

\subsection{Mixed Task-Guided Feature Enhancement}

Currently, pathology LVLMs~\cite{pathasst,quiltllava,pathchat,pallava} commonly use the vision encoder from Contrastive Language-Image Pre-Training (CLIP)~\cite{clip} to convert images into embedding tokens. Although this vision encoder enables alignment between image features and the text space, facilitating the LLM's understanding of image content, it primarily relies on pre-training data consisting of image-caption pairs. This leads the vision encoder to excel at extracting global features of images but limits its ability to perceive local details and spatial structures. However, in pathological diagnosis, accurately identifying foundational pathological concepts and their spatial relationships within pathology WSIs is essential. For instance, in diagnosing microvascular invasion~\cite{Zhang_2023_CVPR}, a pathologist needs to first locate the cancerous region and then inspect surrounding vessels for the presence of cancer cell nuclei, requiring the model to have a nuanced understanding of pathological concepts such as cancerous tissue, blood vessels, normal cell nuclei, and cancer cell nuclei, along with their spatial relationships. Current pathology LVLMs still exhibit limitations in this regard. To address this issue, we focus on both training data and model architecture to enhance the model’s capability in extracting and understanding detailed features.

In terms of training data, we designed a hierarchical instruction fine-tuning dataset covering diverse tasks such as referring expression detection and segmentation, to enhance the pathological feature extraction ability of the visual encoder $\mathcal{V}$ and the visual feature comprehension ability of the LLM $\mathcal{M}$. This dataset systematically constructs concept recognition tasks at three levels: tissue, structure, and cellular, from macro to micro perspectives. 

At the tissue level, tasks include detecting and segmenting cancerous regions in WSI thumbnails, as well as detecting lymph nodes. At the structural level, the focus is on detecting and segmenting blood vessels, bile ducts, and nerves, along with recognizing microvascular invasion, neural invasion, and lymph node invasion. At the cellular level, in addition to basic nucleus classification and detection, more complex tasks requiring inferential abilities were designed, such as “detecting cancer cell nuclei within vessels.” 

To further improve the model's ability to recognize multidimensional pathological features, we also constructed a cross-scale instruction task set that includes organ recognition, cancer subtype identification and grading, tissue type recognition, microsatellite instability detection, and tumor-infiltrating lymphocyte recognition. These self-constructed datasets, combined with publicly available pathology visual question-answering datasets, have formed a training dataset containing 21 types of organs and approximately 490K training samples, significantly enhancing the model's ability to extract pathological features at multiple scales and granularities. 
For details on dataset sources and construction methods, please refer to the \cref{supsec:dataset}.

\begin{figure*}[t!]
    \centering
    \includegraphics[width=\textwidth]{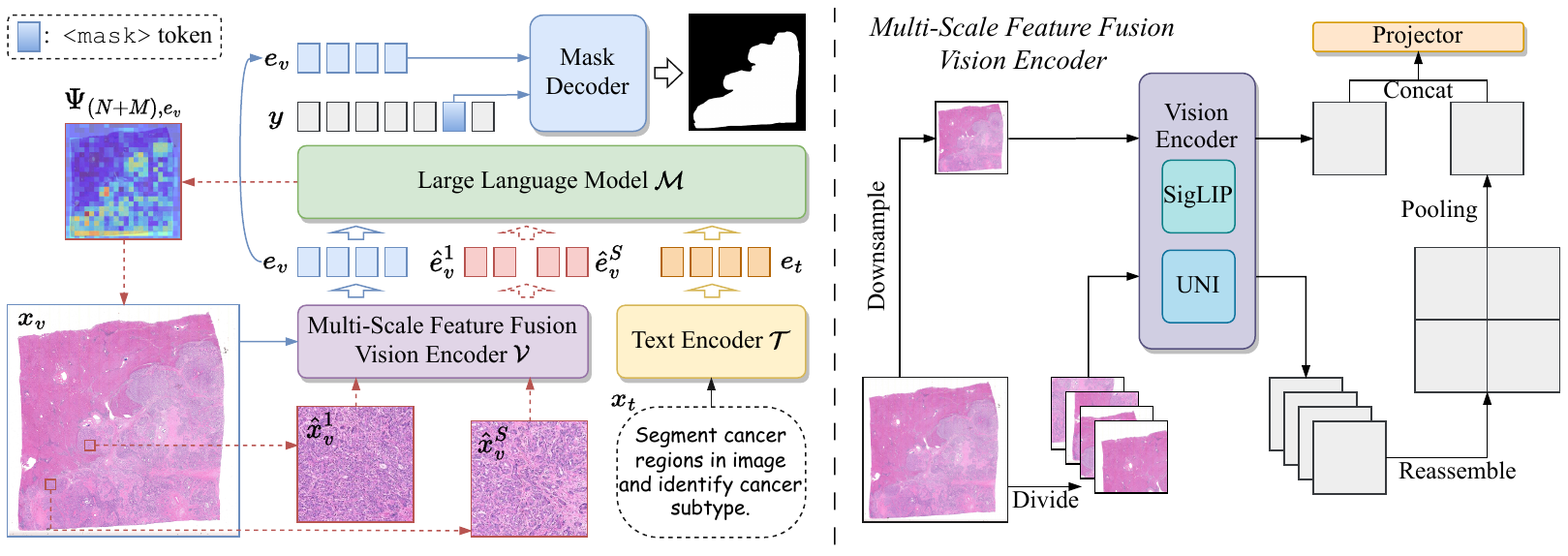}
    \vspace{-1.3em}
    \caption{Overview of the proposed OmniPath. Left: the architecture of OmniPath with the MGFE and PGFC strategy.
    The MGFE model module improvements include a multi-scale feature fusion vision encoder and an additional mask decoder. The PGFC process, shown by the red dashed line, involves inputting a WSI thumbnail and its corresponding prompt into OmniPath. The top-$S$ patches with the highest attention values are selected, and their higher-resolution images are retrieved from the original WSI and added as supplementary input to OmniPath.
    Right: the detailed structure of the multi-scale feature fusion vision encoder.}
    \label{fig:framework}
    \vspace{-1em}
\end{figure*}

In terms of model architecture, we implemented three primary improvements (as shown in \cref{fig:framework}). First, we added an extra ViT named UNI~\cite{UNI}. This ViT, pretrained on large-scale pathology images using the DINOv2~\cite{dinov2} framework, provides critical fine-grained pathological visual features for detection and segmentation tasks. Second, we adopted a multi-scale feature fusion strategy, enabling the model to handle higher-resolution input images without retraining the vision encoder. Specifically, we set a series of increasing input resolutions based on the original resolution supported by the vision encoder, with other resolutions as integer multiples of this base. For the original resolution images, features are directly extracted via the vision encoder. For higher-resolution images, they are divided into image tiles of the original size, each processed separately for feature extraction, and then reassembled based on spatial position. Finally, the features are averaged to match the original feature map dimensions and concatenated along the channel dimension. This approach allows for richer detail extraction without increasing the number of input image tokens $e_v$ to the large language model $\mathcal{M}$.

Third, we introduced a mask encoder and decoder module and added a new \texttt{<mask>} token to the LLM vocabulary to represent segmentation results. The mask encoder encodes binary segmentation maps into embeddings to replace the corresponding \texttt{<mask>} positions in the input, while the mask decoder generates segmentation results based on the image embeddings $e_v$ and the output embedding corresponding to \texttt{<mask>}. The segmentation is optimized using per-pixel binary cross-entropy (BCE) loss and Dice loss. Notably, we attempted to add a dedicated bounding box encoder and decoder for detection tasks but found that this design reduced performance on dense detection tasks, such as nucleus detection. Therefore, we ultimately adopted a strategy that outputs bounding box coordinates directly in relative terms. These enhancements significantly improve the model's performance on multi-scale feature extraction and fine-grained pathological visual tasks.

\subsection{Prompt-Guided Detail Feature Completion}
When using pathology LVLMs for WSI diagnostic analysis, only the thumbnail of the WSI can be used as input, resulting in substantial information loss that affects diagnostic accuracy. Our proposed PGFC strategy dynamically completes missing information based on specific task requirements while maintaining inference efficiency, as shown in \cref{fig:framework}. Specifically, we first remove elements corresponding to image background regions from $\Psi_{(N+M), e_v}$ and select the top-$S$ elements with the highest values from the remaining elements, denoting their indices as $\mathbf{I} = \{i_1, i_2, \dots, i_S\}$. Then, using the index set $\mathbf{I}$, we locate the corresponding tile regions in the original WSI and extract high-resolution patches from these regions, denoted as $\hat{x}_v = \{\hat{x}_v^1, \hat{x}_v^2, \dots, \hat{x}_v^S\}$.

We use $\mathcal{V}$ to extract features for each image in $\hat{x}_v$, obtaining a set of feature sequences $\hat{e}_v = \{\hat{e}_v^1, \hat{e}_v^2, \dots, \hat{e}_v^S\}$, where $\hat{e}_v^s = \mathcal{V}(\hat{x}_v^s)$. In parallel, we encode the positional information of each patch using textual descriptions, denoted as $\hat{x}_t = \{\hat{x}_t^1, \hat{x}_t^2, \dots, \hat{x}_t^S\}$, and obtain the corresponding text embeddings $\hat{e}_t = \{\hat{e}_t^1, \hat{e}_t^2, \dots, \hat{e}_t^S\}$ through $\mathcal{T}$, where $\hat{e}_t^s = \mathcal{T}(\hat{x}_t^s)$. By feeding $\hat{e}_v$ and $\hat{e}_t$ along with the original inputs $e_v$ and $e_t$ into $\mathcal{M}$, we can obtain the final diagnostic result $y = \mathcal{M}(e_v, e_t, \hat{e}_v, \hat{e}_t)$. To mitigate the impact of the large number of $\hat{e}_v$ tokens on $\mathcal{M}$’s inference efficiency, we use the average pooling on each element of $\hat{e}_v$ to reduce the token count of it.

% ############ Experiments ############
\section{Experiments}
\label{sec:expe}

In this section, we first provide the implementation details of OmniPath, followed by comparative results across multiple tasks. Finally, we conduct ablation studies on the key components of OmniPath.

\subsection{Implementation Details}
Based on the pretrained LLaVA-1.5~\cite{llava}, we constructed OmniPath by replacing its CLIP ViT-L 336px~\cite{clip} visual encoder with SigLIP ViT-SO 384px~\cite{siglip} and integrating UNI~\cite{UNI} as an auxiliary vision encoder. The projector utilizes a two-layer MLP with GELU activation. The mask encoder is implemented with ResNet-18~\cite{resnet}, while the mask decoder follows SAM's~\cite{sam} decoder architecture (randomly initialized instead of using pre-trained weights) but directly uses the LVLM's vision encoder described above to replace SAM's original image encoder for feature extraction. During training, all modules of OmniPath participate in end-to-end training with no parameter freezing. We created a multitask dataset encompassing 21 organs and approximately 490K samples for model training (see \cref{supsec:dataset} for detailed data sources and construction methods). Unlike the two-stage training strategy commonly adopted by existing LVLMs, OmniPath requires only a single-stage fine-tuning: trained for 2 epochs on 8 NVIDIA A100 GPUs using the AdamW optimizer, with a learning rate of 2e-5 and a global batch size of 128. We set $S=8$, and compare OmniPath with LLaVA-1.5~\cite{llava}, LLaVA-Med~\cite{llavamed}, Quilt-LLaVA~\cite{quiltllava}, and PA-LLaVA~\cite{pallava}.

For a fair comparison, we fine-tuned Quilt-LLaVA using the same training dataset and hyperparameters. The test results are shown in the table below as ``Quilt-LLaVA (FT)''. Since its model architecture is identical to that of LLaVA, with only the weights differing, this result partially reflects the fine-tuning performance of other LLaVA-based methods.

\subsection{Comparison on Pathology Diagnostic Tasks}
To evaluate the pathology diagnostic performance of various LVLMs, we conducted a series of clinically relevant pathology diagnostic experiments, divided by diagnostic granularity into patch-level and slide-level categories. The patch-level experiments included subtyping and grading tasks for a range of common cancers, such as hepatocellular carcinoma subtyping (HCC-S) and grading (HCC-G), intrahepatic cholangiocarcinoma subtyping (ICC-S) and grading (ICC-G), renal cell carcinoma subtyping (RCC-S), lung cancer subtyping (LUNG-S) and grading (LUNG-G), gastric adenocarcinoma Lauren subtyping (STAD-L) and grading (STAD-G). In addition, other tasks related to pathology concept recognition or diagnosis included: microvascular invasion identification (MVI), neural invasion identification (NI), pan-cancer identification across 32 types (PanCancer), organ classification (OC), tissue classification (TC), tumor-infiltrating lymphocyte identification (TIL), microsatellite instability detection in colorectal cancer (MSI), and seven-class gastric lesion recognition (GLR-7). The accuracy of each model on these tasks is detailed in \cref{tab:1}.

Slide-level experiments included not only the same subtyping and grading tasks as the patch-level but also additional tasks, such as lymph node metastasis diagnosis (LNM), HCC prognosis prediction (HCC-P), and colorectal cancer prognosis prediction (CRC-P), using WSI thumbnails as image input. The accuracy of each model on slide-level tasks is presented in \cref{tab:2}.

It can be observed that OmniPath achieves the best performance across all patch-level and slide-level pathological diagnosis tasks. For most cancer subtype classification and grading tasks, OmniPath achieves accuracy rates exceeding 70\%, and in many cases, surpassing 90\%. In contrast, the accuracy rates of other models are generally below 70\%. This demonstrates that OmniPath is more suitable for clinical applications to assist pathologists in diagnosis. Furthermore, OmniPath also exhibits strong recognition capabilities for features such as microvascular invasion, neural invasion, and tumor-infiltrating lymphocytes. This significantly reduces the extensive effort required by pathologists to meticulously examine detailed pathological lesions during slide review.
Moreover, OmniPath outperforms Quilt-LLaVA (FT), demonstrating the effectiveness of the MGFE and PGFC strategies.

\begin{table*}[!t]
\centering
\caption{Accuracy (\%) comparison on patch-level pathology diagnostic tasks.}
\label{tab:1}
\vspace{-0.3em}
\resizebox{\textwidth}{!}{
\begin{tabular}{l|ccccccccc|cccccccc}
\hline
Method      & HCC-S          & HCC-G          & ICC-S          & ICC-G          & RCC-S          & LUNG-S         & LUNG-G         & STAD-L         & STAD-G         & MVI            & NI             & PanCancer      & OC             & TC             & TIL            & MSI            & GLR-7          \\ \hline
LLaVA-1.5   & 25.76          & 0.16           & 16.37          & 3.97           & 23.82          & 21.03          & 14.66          & 15.43          & 9.32           & 18.15          & 55.07          & 3.17           & 3.81           & 9.89           & 51.92          & 50.02          & 13.93          \\
LLaVA-Med   & 31.23          & 7.78           & 38.56          & 16.45          & 52.34          & 38.77          & 36.45          & 28.12          & 25.23          & 16.89          & 54.67          & 28.34          & 21.29          & 33.88          & 58.42          & 47.21          & 39.87          \\
PA-LLaVA    & 82.45          & 40.56          & 66.78          & 79.34          & 73.67          & 76.78          & 60.23          & 64.89          & 59.89          & 61.56          & 67.12          & 46.78          & 48.23          & 68.45          & 76.34          & 62.45          & 58.67          \\
Quilt-LLaVA & 78.34          & 42.23          & 68.45          & 81.56          & 75.23          & 83.12          & 57.89          & 62.89          & 61.23          & 50.78          & 79.45          & 44.56          & 49.67          & 70.56          & 77.34          & 63.12          & 59.45          \\
Quilt-LLaVA (FT)    & 92.89          & 51.90          & 80.84          & 85.55          & 80.07          & 95.27          & 68.01          & 88.22          & 78.09          & 96.59          & 90.75          & 49.73          & 57.14          & 74.43          & 81.90          & 67.08          & 74.39          \\
\rowcolor[HTML]{EFEFEF} 
OmniPath    & \textbf{97.09} & \textbf{63.07} & \textbf{86.04} & \textbf{96.53} & \textbf{90.17} & \textbf{96.55} & \textbf{71.98} & \textbf{91.56} & \textbf{87.59} & \textbf{97.79} & \textbf{93.83} & \textbf{54.44} & \textbf{69.52} & \textbf{86.66} & \textbf{89.88} & \textbf{73.78} & \textbf{79.01} \\ \hline
\end{tabular}
}
\end{table*}

\begin{table*}[!t]
\centering
\vspace{-0.3em}
\caption{Accuracy (\%) comparison on slide-level pathology diagnostic tasks.}
\label{tab:2}
\vspace{-0.3em}
\resizebox{\textwidth}{!}{
\begin{tabular}{l|ccccccccc|ccc}
\hline
Method      & HCC-S          & HCC-G          & ICC-S          & ICC-G          & RCC-S          & LUNG-S         & LUNG-G         & STAD-L         & STAD-G         & LNM            & HCC-P          & CRC-P          \\
\hline
LLaVA-1.5   & 14.31          & 37.50          & 17.25          & 32.55          & 9.09           & 19.45          & 20.33          & 15.44          & 18.77          & 61.04          & 0.00           & 31.88          \\
LLaVA-Med   & 15.32          & 35.42          & 23.56          & 38.53          & 10.39          & 23.88          & 26.48          & 18.59          & 19.95          & 63.21          & 8.45           & 32.78          \\
PA-LLaVA    & 77.96          & 53.05          & 70.84          & 74.45          & 73.75          & 74.74          & 70.44          & 43.77          & 55.25          & 55.70          & 52.63          & 69.64          \\
Quilt-LLaVA & 77.73          & 57.88          & 67.47          & 83.69          & 66.78          & 70.96          & 66.14          & 39.12          & 53.90          & 64.39          & 47.38          & 65.71          \\
Quilt-LLaVA (FT) & 89.74          & 64.23          & 82.35          & 93.58          & 81.49          & 91.73          & 77.59          & 44.14          & 70.08          & 72.74          & 59.32          & 76.81          \\
\rowcolor[HTML]{EFEFEF} 
OmniPath    & \textbf{98.40} & \textbf{70.83} & \textbf{87.76} & \textbf{99.08} & \textbf{87.88} & \textbf{98.72} & \textbf{83.09} & \textbf{52.72} & \textbf{76.69} & \textbf{79.22} & \textbf{66.10} & \textbf{85.51} \\
\hline
\end{tabular}
}
\end{table*}

\subsection{Comparison on Zero-Shot Classification Tasks}
To evaluate the clinical generalization capability of OmniPath, we employed a zero-shot classification paradigm, testing on several widely-used academic pathology datasets that were not included in the training set. The evaluation covered two levels: patch-level tasks using the CCRCC~\cite{CCRCC}, MHIST~\cite{MHIST}, and NCT-CRC~\cite{NCT-CRC} datasets, and slide-level tasks using the PANDA~\cite{PANDA}, DHMC~\cite{DHMC}, and CAMELYON17~\cite{Camelyon17} datasets. Using a closed-ended question-answering approach, the model was required to classify images into predefined categories specific to each dataset. The performance comparison of all models on these zero-shot test sets is presented in \cref{tab:3}.

It is shown that OmniPath consistently outperforms other models across both patch-level and slide-level datasets in zero-shot classification tasks, highlighting its strong generalization ability in pathological image analysis. Notably, on the PANDA and CAMELYON17 slide-level datasets, OmniPath achieved the highest accuracy rates of 79.15\% and 59.33\%, respectively, which significantly surpasses the performance of other models. This superior performance in zero-shot classification indicates OmniPath’s robustness in handling diverse pathological image data and reinforces its potential for clinical applications where labeled training data may be limited.

\begin{table}[t]
\centering
\begin{minipage}{0.49\textwidth}
\centering
% 表zero-shot
\caption{Accuracy (\%) on zero-shot classification.}
\vspace{0.35em}
\label{tab:3}
\resizebox{\textwidth}{!}{
\centering
\begin{tabular}{l|ccc|ccc}
\hline
                         & \multicolumn{3}{c|}{Patch-Level Dataset}         & \multicolumn{3}{c}{Slide-Level Dataset}          \\
\multirow{-2}{*}{Method} & CCRCC          & MHIST          & NCT-CRC        & PANDA          & DHMC           & CAMELYON17     \\ \hline
LLaVA-1.5                & 15.47          & 69.31          & 11.58          & 62.43          & 31.84          & 19.44          \\
LLaVA-Med                & 13.67          & 33.82          & 12.71          & 57.40          & 10.57          & 39.03          \\
PA-LLaVA                 & 15.43          & \textbf{70.02} & 15.19          & 70.33          & 27.88          & 34.45          \\
Quilt-LLaVA              & 17.18          & 68.46          & 13.84          & 64.06          & 28.48          & 48.27          \\
\rowcolor[HTML]{EFEFEF} 
OmniPath                 & \textbf{23.67} & 69.77          & \textbf{20.67} & \textbf{79.15} & \textbf{34.62} & \textbf{59.33} \\ \hline
\end{tabular}
}
\end{minipage}%
\hfill
\begin{minipage}{0.5\textwidth}
\centering
%表seg
\caption{Performance comparison on referring segmentation tasks. Dice coefficient (\%) is used as the evaluation metric.}
\label{tab:5}
\resizebox{\textwidth}{!}{
\begin{tabular}{l|cccccc|ccc}
\hline
\multirow{2}{*}{Method} & \multicolumn{6}{c|}{Slide-Level Cancer Region}            & \multicolumn{3}{c}{Tissue Structure} \\
                        & HCC   & ICC   & RCC   & GBM   & LUAD  & BC    & NS         & NIS        & LNMS       \\ \hline
Quilt-LLaVA (FT)        & 62.10 & 11.95 & 23.82 & 4.82 & 18.67 & 16.77 & 24.32      & 3.25      & 4.01      \\
\rowcolor[HTML]{EFEFEF}OmniPath & 95.51 & 93.60 & 85.11 & 50.01 & 93.03 & 74.64 & 89.23      & 62.14      & 56.94      \\ \hline
\end{tabular}
}
\end{minipage}
\end{table}

\begin{table*}[!t]
\centering
\vspace{-0.3em}
\caption{Performance comparison on pathology referring detection tasks.}
\label{tab:4}
\vspace{-0.5em}
\resizebox{\textwidth}{!}{
\begin{tabular}{l|l|cccccc|ccc|ccc}
\hline
\multirow{2}{*}{Method} & \multirow{2}{*}{Metric} & \multicolumn{6}{c|}{Slide-Level Cancer Region}             & \multicolumn{3}{c|}{Tissue Structure} & \multicolumn{3}{c}{Cell Nucleus} \\
                        &                         & HCC   & ICC    & RCC   & GBM   & LUAD  & BC    & LD          & VD         & ND         & MoNuSeg    & NuCLS   & PanNuke   \\ \hline
\multirow{2}{*}{Quilt-LLaVA (FT)} 
& F1-Score (\%) & 84.64 & 77.27 & 63.33 & 14.38 & 63.33 & 11.76 & 32.76       & 69.81 & 15.03 & 14.27 & 5.63 & 10.17     \\
& IoU (\%) & 78.14 & 78.62  & 73.38 & 63.69 & 69.86 & 87.26 & 67.03 & 69.03      & 66.62 & 63.05 & 19.59 & 26.73     \\ 
\rowcolor[HTML]{EFEFEF} & F1-Score (\%)               & 92.13 & 100.00 & 90.91 & 40.85 & 90.91 & 41.38 & 82.70       & 89.52      & 72.50      & 83.98      & 34.30   & 45.30     \\
\rowcolor[HTML]{EFEFEF}\multirow{-2}{*}{OmniPath} & IoU (\%)                     & 91.97 & 91.85  & 89.59 & 73.85 & 87.96 & 72.75 & 87.34       & 83.00      & 79.75      & 79.65      & 44.98   & 59.99     \\
\hline
\end{tabular}
\vspace{-0.5em}
}
\end{table*}

\subsection{Detection and Segmentation Performance}
In pathological diagnosis, detection and segmentation tasks are as critical as classification tasks.
Since existing pathology LVLMs lack detection and segmentation capabilities, we only compare with Quilt-LLaVA (FT). To enable Quilt-LLaVA to perform detection and segmentation, we used the same training dataset as OmniPath and converted the segmentation masks in the dataset into corresponding polygons with up to 50 vertices each. We then had Quilt-LLaVA generate the vertex coordinate sequences of the polygons as text sequences.
The detection tasks cover various cancer region identifications, including HCC, ICC, RCC, glioblastoma (GBM), lung adenocarcinoma (LUAD), and bladder cancer (BC). Additionally, it involves detecting tissue structures such as lymph nodes (LD), vessels (VD), and nerves (ND), as well as cell nuclei detection in datasets like MoNuSeg~\cite{MoNuSeg} (without categories), NuCLS~\cite{NuCLS}, and PanNuke~\cite{PanNuke} (with categories). The segmentation tasks not only include the segmentation of cancer regions covered by the detection tasks but also nerve segmentation (NS), nerve invasion region segmentation (NIS), and lymph node metastasis segmentation (LNMS). Detection tasks are evaluated using F1-score and IoU, while segmentation tasks are assessed using the Dice coefficient.
\cref{tab:5} and \cref{tab:4} show the segmentation and detection performance,  respectively.

The ``Cancer Region'' detection and segmentation tasks are slide-level, and all remaining tasks are patch-level. OmniPath outperforms Quilt-LLaVA (FT) in both detection and segmentation.
While OmniPath’s performance may not surpass specialized smaller models on certain tasks (see \cref{supsec:vision-model}), it offers a unique advantage in inferential capabilities. As shown in \cref{fig:sample}, OmniPath can accurately detect cancer cell nuclei within blood vessels according to instructions (see \cref{supsec:closed} for quantitative results)—an ability that current specialized small models find challenging to achieve.

\subsection{Ablation Study}

\begin{wrapfigure}{r}{0.35\textwidth}
\vspace{-3em}
\begin{minipage}{0.35\textwidth}
\centering
\captionof{table}{Ablation study of PGFC strategy on the slide-level diagnostic tasks. ``Random'' refers to selecting $\hat{x}_v$ at random, rather than selecting based on $\mathbf{I}$.}
\label{tab:6}
\resizebox{\textwidth}{!}{
\centering
\begin{tabular}{l|ccc}
\hline
Task   & w/o PGFC & Random & w/ PGFC        \\ \hline
HCC-S  & 20.14    & 79.46  & \textbf{98.40} \\
HCC-G  & 52.84    & 45.83  & \textbf{70.83} \\
ICC-S  & 85.71    & 85.71  & \textbf{87.76} \\
ICC-G  & 66.37    & 83.72  & \textbf{99.08} \\
RCC-S  & 60.61    & 75.76  & \textbf{87.88} \\
LUNG-S & 96.58    & 96.43  & \textbf{98.72} \\
LUNG-G & 65.54    & 63.21  & \textbf{83.09} \\
STAD-L & 44.60    & 35.07  & \textbf{52.72} \\
STAD-G & 72.92    & 75.31  & \textbf{76.69} \\ \hline
Average    & 62.81    & 71.17  & \textbf{83.91} \\ \hline
\end{tabular}
}
\end{minipage}
\end{wrapfigure}

The ablation study in \cref{tab:6} underscores the effectiveness of the PGFC strategy in enhancing OmniPath’s performance across slide-level diagnostic tasks. Three configurations were tested: (1) without PGFC, (2) replacing the top-S elements in $\Psi_{(N+M), e_v}$ with random selection of $\hat{x}_v$, and (3) with the designed PGFC strategy. Removing PGFC resulted in an average accuracy drop of 21.1\%, while random selection led to a decrease of 12.7\% compared to using PGFC. Notably, for tasks like HCC-S and ICC-G, PGFC boosted accuracy significantly, demonstrating its ability to enhance model focus on essential features, which is critical for accurate diagnosis across complex pathology tasks. In certain tasks, such as HCC-G and STAD-L, random selection results in performance that is even lower than without PGFC, indicating that incorrectly supplementing detailed features can also impair model performance. The ablation study of MGFE model module improvements is in \cref{supsec:abofMGFE}.

% ############ Conclusion and Limitations ############
\section{Conclusion}
\label{sec:con}

This paper introduces OmniPath, a pathology-focused LVLM, fundamentally shaped by two innovative strategies addressing key limitations in existing models. The mixed task-guided feature enhancement strategy enables precise extraction of lesion-specific details, which is crucial for accurate diagnostic assessments. Meanwhile, the prompt-guided detail feature completion strategy combines coarse global context with fine-grained detail in response to clinical needs. Together, these strategies allow OmniPath to achieve a comprehensive and balanced feature extraction, validated across a wide range of pathology tasks. These advancements underscore OmniPath’s potential as a transformative tool in digital pathology.

\textbf{Limitations and Future Work.}
OmniPath currently faces three main limitations: First, the model lacks sufficient depth in medical and pathological expertise, primarily due to training data being dominated by image-caption pairs, with limited integration of cutting-edge pathology knowledge and literature. To address this, we plan to enhance the model’s knowledge base using retrieval-augmented generation (RAG) techniques. Second, its performance in zero-shot classification tasks needs improvement. We will introduce a multi-agent framework to enable specialized agents to assist in making more accurate diagnostic decisions for challenging cases. Finally, the model's reasoning capability requires enhancement, as it has not yet reached the level for independent diagnosis. To resolve this, we will collect pathologists’ diagnostic process data and integrate it with reinforcement learning to improve reasoning, aiming for autonomous diagnosis and reduced physician workload.

% ################# Acknowledgments Part #################
% \begin{ack}
% Use unnumbered first level headings for the acknowledgments. All acknowledgments
% go at the end of the paper before the list of references. Moreover, you are required to declare
% funding (financial activities supporting the submitted work) and competing interests (related financial activities outside the submitted work).
% More information about this disclosure can be found at: \url{https://neurips.cc/Conferences/2025/PaperInformation/FundingDisclosure}.

% Do {\bf not} include this section in the anonymized submission, only in the final paper. You can use the \texttt{ack} environment provided in the style file to automatically hide this section in the anonymized submission.
% \end{ack}

% ################# Begin of References #################
%\section*{References}
% {
%     \small
%     \bibliographystyle{ieeenat_fullname}
%     \bibliography{main}
% }
\printbibliography[title={References}, heading=bibintoc]

%%%%%%%%%%%%%%%%%%%%%%%%%%%%%% Appendix %%%%%%%%%%%%%%%%%%%%%%%%%%%%%%

\clearpage
\appendix

\newrefsection % 关键！新建一个独立的参考文献区块

{
\centering
\Large
\textbf{Efficient and Comprehensive Feature Extraction in \\
Large Vision-Language Model for Pathology Analysis}\\
\vspace{0.5em}
Appendix \\
}

\begin{figure}[H]
\centering
\includegraphics[width=\textwidth]{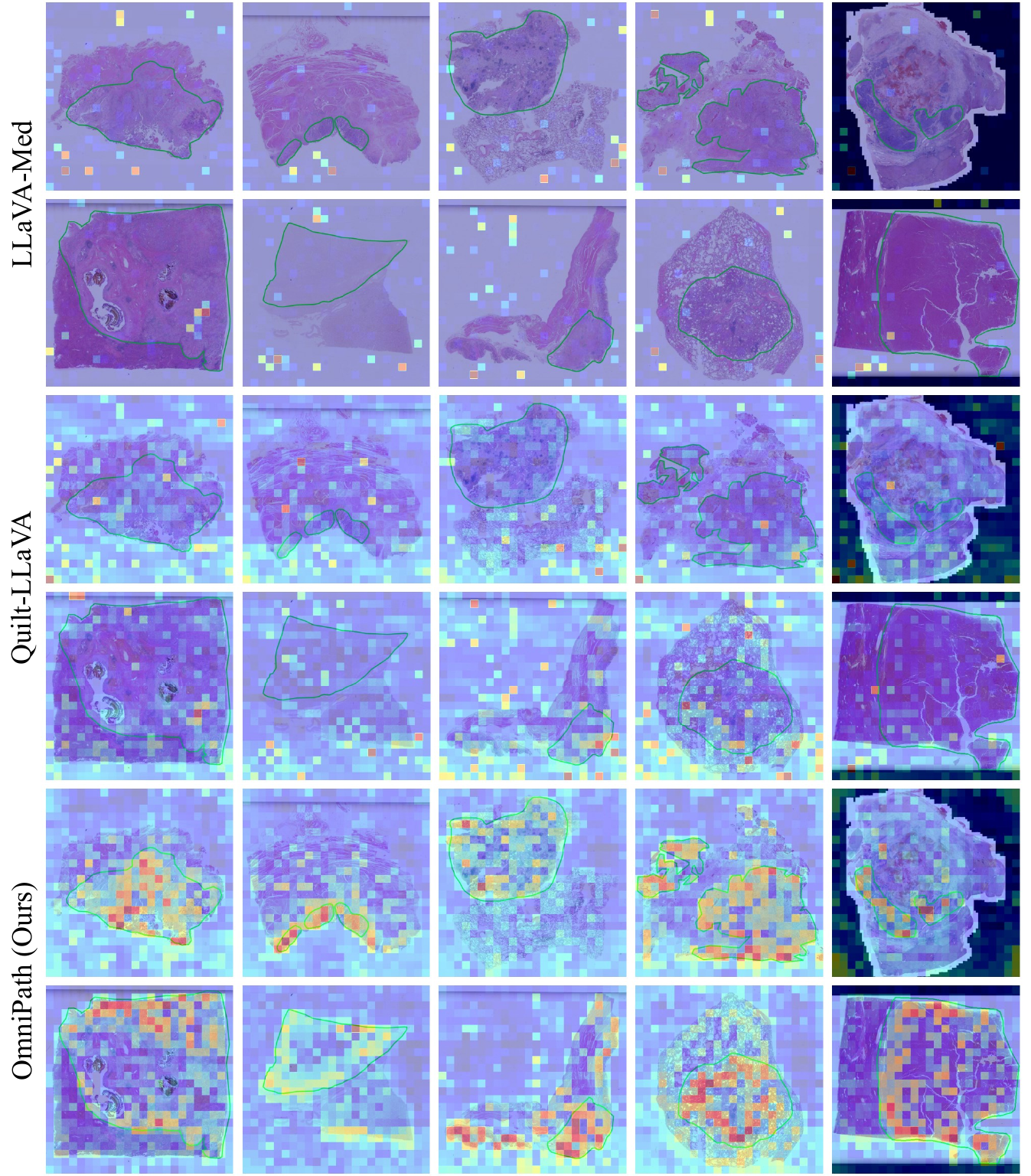}
\caption{
More samples of the attention of upcoming token on image tokens (like the fist column in \cref{fig:attn} of the main paper).
The intensity of attention values is mapped from blue (low) to red (high), and the green contours on the pathology slides mark cancerous regions annotated by pathologists.
It can be observed that the image tokens focused on by OmniPath are generally concentrated within the cancerous regions.
}
\label{supfig:sample}
\end{figure}

To facilitate a better understanding of the value and significance of this work, as well as to thoroughly demonstrate the effectiveness and applicability of the OmniPath in addressing diverse pathology-related tasks, we have supplemented more visualization results, more experiment results, and data sources and construction methods, as follows.

\section{More Visualization Results of Decision-Dependent Image Tokens}

To demonstrate the generalizability of the analysis conducted in \cref{sec:heatmap_analysis}, we visualized $\Psi_{(N+M), e_v}$ as heatmaps on additional pathology WSIs, as shown in \cref{supfig:sample}. These WSIs include samples from various cancers such as hepatocellular carcinoma, lung cancer, gastric cancer, renal cancer, bladder cancer, prostate cancer, and glioblastoma, none of which were involved in model training. The prompt used for visualization was \textit{``What cancer subtype is shown in this image?''}.

In the The heatmaps generated by LLaVA-Med~\cite{llavamed} and Quilt-LLaVA~\cite{quiltllava} in the \cref{supfig:sample} show that image tokens with high attention weights are distributed rather randomly, with most falling outside the cancerous regions and even outside tissue regions. This distribution is clearly inconsistent with the areas that need attention for cancer subtype identification. In contrast, the tokens focused on by OmniPath are predominantly distributed within cancerous regions annotated by pathologists. This indicates that the optimized OmniPath can more accurately capture critical diagnostic features in pathology WSIs, thereby performing user-directed tasks more effectively.

\section{t-SNE Visualization Results of Learnt Image Features}

To demonstrate the benefits brought by the MGFE strategy—specifically, the use of mixed-task data and the enhancements made to the visual encoder—we visualized the image features extracted by the visual encoder using t-SNE in a low-dimensional space. Specifically, we extracted image features using both Quilt-LLaVA~\cite{quiltllava} and our OmniPath on the slide-level test sets of HCC, ICC, and RCC. Based on the annotated cancer regions, we categorized features outside the cancer areas as ``Benig'' and those inside as ``Cancer'', and then performed t-SNE visualization on these two classes of features.

\begin{figure}[h]
\centering
\includegraphics[width=\textwidth]{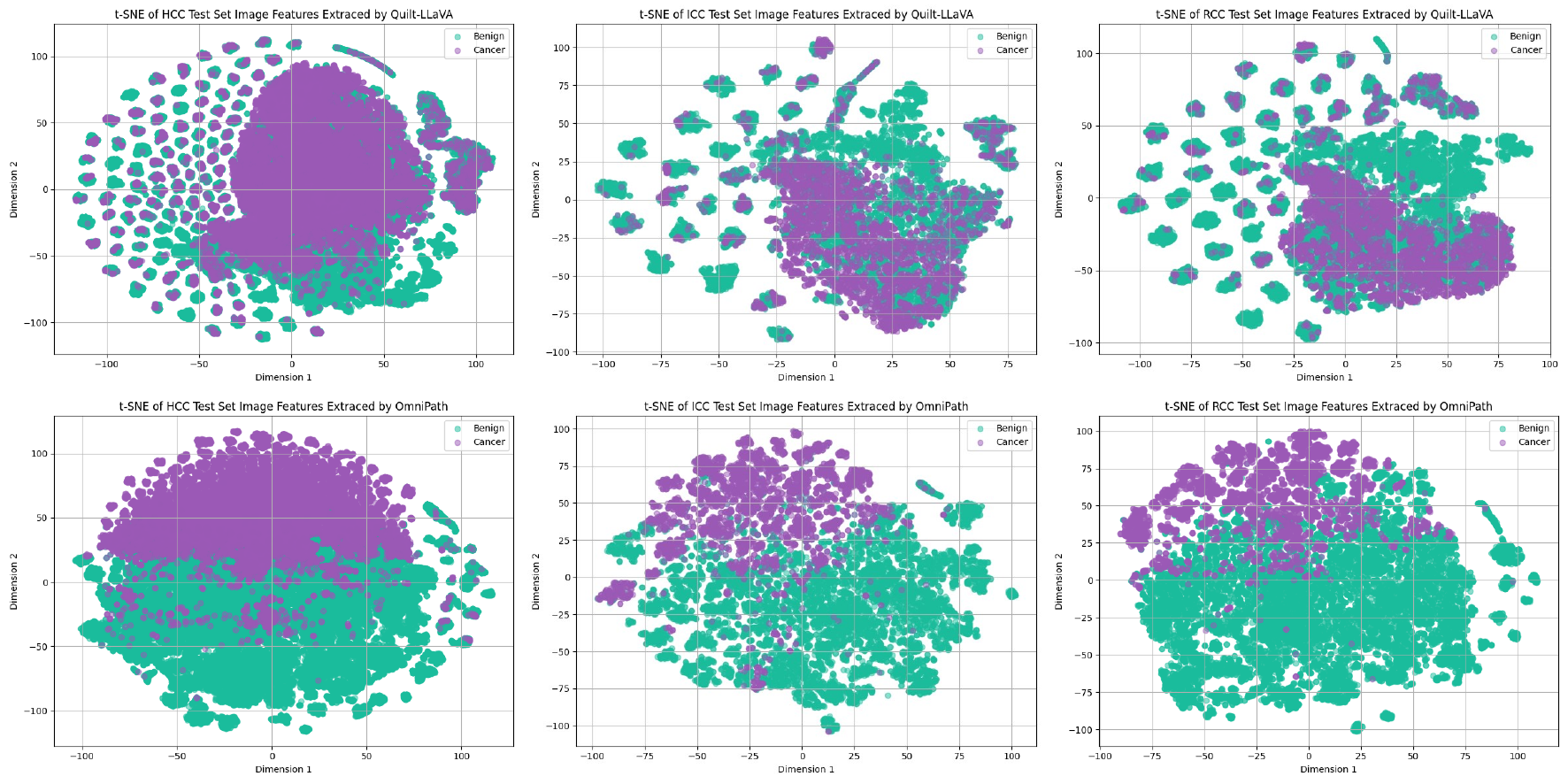}
\caption{
t-SNE visualization results of slide-level image features extracted by vision encoders of Quilt-LLaVA (first row) and OmniPath (second row), respectively. Based on the cancer regions annotated by pathologists, we classify the image feature tokens into two categories: benign and cancer. It can be seen that the image features extracted by OmniPath demonstrate superior inter-class discriminability and intra-class diversity.
}
\label{supfig:tsne}
\end{figure}

As shown in the \cref{supfig:tsne}, compared to Quilt-LLaVA~\cite{quiltllava} (first row), the features learned by OmniPath (second row) exhibit better separation between the two classes, indicating that OmniPath has captured more task-relevant and discriminative features. Moreover, the feature distribution within the same class is more uniform in OmniPath, suggesting stronger representational capacity.

\section{Comparison with Traditional Vision Models on Detection and Segmentation Tasks}
\label{supsec:vision-model}

OmniPath can perform detection and segmentation tasks that other pathological LVLMs (e.g., Quilt-LLaVA~\cite{quiltllava}) cannot accomplish. To further validate its utility in these tasks, we compared OmniPath with traditional vision models specialized in either detection or segmentation. Specifically, we selected representative and high-performing detection and segmentation models—YOLO11~\cite{yolo11} and nnU-Net V2~\cite{nnunet}—and trained and evaluated them on selected detection and segmentation tasks using their default hyperparameters. The evaluation metrics for detection and segmentation are F1-score and Dice, respectively. OmniPath and traditional models have mixed results across datasets. On average, traditional model (70.08) performs better than OmniPath (68.05) in detection, while OmniPath (77.80) outperforms traditional model (76.95) in segmentation. The detailed results are in \cref{suptab:vision-det,suptab:vision-seg}, respectively.

OmniPath's performance currently doesn't surpass traditional visual models, mainly because traditional models are specifically optimized for their tasks. Additionally, there is often a need to detect numerous small targets in pathology detection (e.g., nuclei), whereas current LVLM mainly detect fewer, larger targets. 

However, the use of LLM offers OmniPath several advantages that traditional visual models lack: First, it can handle tasks requiring reasoning, such as ``detecting cancer cell nuclei within vessels'', which traditional models cannot do. A nuclear detection model cannot determine if a nucleus is inside a vessel. Second, OmniPath can address various data types for different tasks, offering better generalization than traditional models. Third, if expert-level classification, detection, or segmentation is needed, traditional models can be pre-trained for specific tasks and integrated with OmniPath as callable tools, allowing dynamic invocation based on user needs, thus enhancing model capabilities without retraining.

\begin{table}[h]
\centering
\caption{Comparison results between OmniPath and traditional object detection model across multiple datasets. The metric used in the table is F1-score. As seen, OmniPath and traditional model each have their strengths and weaknesses on different datasets. On average, the traditional object detection model performs better.}
\label{suptab:vision-det}
\resizebox{0.8\textwidth}{!}{
\begin{tabular}{llllllll}
\toprule
Detection Model & LD    & VD    & ND    & MoNuSeg & NuCLS & PanNuke & Average \\
\midrule
Yolo11m   & 85.93 & 84.69 & 72.39 & 84.21   & 31.21 & 62.05   & 70.08   \\
OmniPath  & 82.70 & 89.52 & 72.50 & 83.98   & 34.30 & 45.30   & 68.05   \\
\bottomrule
\end{tabular}
}
\end{table}

\begin{table}[h]
\centering
\caption{Comparison results between OmniPath and traditional semantic segmentation model across multiple datasets. The metric used in the table is Dice. As shown, OmniPath and traditional model each have their strengths and weaknesses on different datasets. On average, OmniPath performs better.}
\label{suptab:vision-seg}
\resizebox{0.97\textwidth}{!}{
\begin{tabular}{lllllllllll}
\toprule
Segmentation Model & HCC   & ICC   & RCC   & GBM   & LUAD  & BC    & NS    & NIS   & LNMS  & Average \\
\midrule
nnU-Net V2   & 94.30 & 92.86 & 90.89 & 52.27 & 91.64 & 80.82 & 67.56 & 68.74 & 53.43 & 76.95   \\
OmniPath     & 95.51 & 93.60 & 85.11 & 50.01 & 93.03 & 74.64 & 89.23 & 62.14 & 56.94 & 77.80   \\
\bottomrule
\end{tabular}
}
\end{table}

\section{Ablation Study of MGFE Model Module Improvements}
\label{supsec:abofMGFE}

The model module improvements in MGFE are primarily designed to enhance the model’s capability in feature extraction and understanding for individual images. Therefore, we conducted ablation studies on patch-level tasks to evaluate their effectiveness. Specifically, we individually removed the additional ViT UNI and the multi-scale feature fusion (MSFF) strategy, and finally removed both components to assess the impact of each module on overall performance. The resulting model architecture, after removing both components, is essentially similar to the original LLaVA.

The ablation results for patch-level diagnostic tasks are shown in \cref{suptab:mgfe-cls}. Both UNI and MSFF contribute to improved performance, as removing either component results in a performance drop. Among the two, the removal of UNI leads to a more pronounced degradation. This is because UNI, trained with the DINOv2 paradigm, is more effective at capturing the fine-grained features critical for pathological diagnosis, whereas SigLIP, trained via vision-language contrastive learning, tends to overlook such details.

The ablation results for patch-level detection tasks are presented in \cref{suptab:mgfe-det}. Similarly, both UNI and MSFF contribute significantly to overall performance, but their impact varies across different types of detection tasks. UNI has a more substantial effect on nucleus detection performance, with a notable decline observed when it is removed. In contrast, MSFF provides greater benefits in detecting tissue structures such as lymph nodes, vessels, and nerves. This is because nucleus detection involves smaller and more uniform targets, requiring the model to focus on fine-grained image details, whereas tissue structure detection involves targets with greater size variability, making multi-scale feature representation more crucial.

\begin{table}[h]
\centering
\caption{Ablation study of MGFE model module improvements on the patch-level diagnostic tasks. The metric in the table is accuracy (\%). MSFF is the multi-scale feature fusion module.}
\label{suptab:mgfe-cls}
\resizebox{\textwidth}{!}{
\begin{tabular}{lccccccccc|cccccccc}
\toprule
Method           & HCC-S          & HCC-G          & ICC-S          & ICC-G          & RCC-S          & LUNG-S         & LUNG-G         & STAD-L         & STAD-G         & MVI            & NI             & PanCancer      & OC             & TC             & TIL            & MSI            & GLR-7          \\ \midrule
w/o UNI and MSFF & 91.07          & 50.71          & 78.15          & 82.79          & 81.02          & 93.38          & 68.23          & 85.52          & 77.62          & 94.88          & 90.32          & 35.83          & 61.76          & 74.33          & 82.69          & 64.65          & 72.51          \\
w/o UNI          & 94.47          & 55.39          & 81.95          & 87.05          & 85.61          & 95.23          & 70.73          & 88.68          & 85.79          & 96.79          & 92.77          & 39.88          & 62.65          & 75.09          & 82.33          & 70.51          & 74.16          \\
w/o MSFF         & 96.53          & 62.10          & 85.51          & 95.86          & 83.44          & 95.77          & 71.88          & 89.49          & 82.91          & 95.29          & 90.54          & 47.17          & 68.57          & 80.27          & 86.98          & 65.60          & 78.93          \\
OmniPath         & \textbf{97.09} & \textbf{63.07} & \textbf{86.04} & \textbf{96.53} & \textbf{90.17} & \textbf{96.55} & \textbf{71.98} & \textbf{91.56} & \textbf{87.59} & \textbf{97.79} & \textbf{93.83} & \textbf{54.44} & \textbf{69.52} & \textbf{86.66} & \textbf{89.88} & \textbf{73.78} & \textbf{79.01} \\
\bottomrule
\end{tabular}
}
\end{table}

\begin{table}[h]
\centering
\caption{Ablation study of MGFE model module improvements on the patch-level detection tasks.}
\label{suptab:mgfe-det}
\resizebox{\textwidth}{!}{
\begin{tabular}{l|cccccc|cccccc}
\toprule
\multirow{3}{*}{Method} & \multicolumn{6}{c|}{Tissue Structure Detection}                                                                                               & \multicolumn{6}{c}{Cell Nucleus Detection}                                                                                                    \\ \cline{2-13} 
                        & \multicolumn{2}{c|}{LD}                              & \multicolumn{2}{c|}{VD}                              & \multicolumn{2}{c|}{ND}         & \multicolumn{2}{c|}{MoNuSeg}                         & \multicolumn{2}{c|}{NuCLS}                           & \multicolumn{2}{c}{PanNuke}     \\
                        & F1             & \multicolumn{1}{c|}{IoU}            & F1             & \multicolumn{1}{c|}{IoU}            & F1             & IoU            & F1             & \multicolumn{1}{c|}{IoU}            & F1             & \multicolumn{1}{c|}{IoU}            & F1             & IoU            \\ \hline
w/o UNI and MSFF        & 77.24          & \multicolumn{1}{c|}{72.58}          & 74.60          & \multicolumn{1}{c|}{75.83}          & 55.92          & 68.31          & 61.67          & \multicolumn{1}{c|}{69.09}          & 25.23          & \multicolumn{1}{c|}{34.55}          & 22.71          & 45.66          \\
w/o UNI                 & 81.33          & \multicolumn{1}{c|}{82.37}          & 81.03          & \multicolumn{1}{c|}{82.65}          & 71.39          & 76.71          & 73.80          & \multicolumn{1}{c|}{74.08}          & 29.10          & \multicolumn{1}{c|}{40.18}          & 33.37          & 48.45          \\
w/o MSFF                & 77.01          & \multicolumn{1}{c|}{84.58}          & 75.93          & \multicolumn{1}{c|}{77.43}          & 68.41          & 77.43          & 82.69          & \multicolumn{1}{c|}{77.66}          & 33.61          & \multicolumn{1}{c|}{44.49}          & 44.38          & 59.97          \\
OmniPath                & \textbf{82.70} & \multicolumn{1}{c|}{\textbf{87.34}} & \textbf{89.52} & \multicolumn{1}{c|}{\textbf{83.00}} & \textbf{72.50} & \textbf{79.75} & \textbf{83.98} & \multicolumn{1}{c|}{\textbf{79.65}} & \textbf{34.30} & \multicolumn{1}{c|}{\textbf{44.98}} & \textbf{45.30} & \textbf{59.99} \\
\bottomrule
\end{tabular}
}
\end{table}

\section{Comparison with SOTA Closed-Source LVLMs}
\label{supsec:closed}

To further verify the effectiveness and superiority of our pathology-specialized LVLM, developed based on our data and methodology, in addressing pathology-related tasks, we compared OmniPath with existing state-of-the-art (SOTA) proprietary LVLMs. Specifically, we selected two widely used and market-validated proprietary LVLMs: ``ChatGPT 4o'' developed by OpenAI and ``Gemini 2.5 Pro'' developed by Google. We submitted images and questions from the test set to each model via their respective APIs and compared the predicted answers with the ground truth to compute evaluation metrics. The specific model versions accessed via API are listed in the comparison table below. Due to budget constraints, we conducted evaluations on a subset of tasks only.

For diagnostic tasks, we selected five slide-level classification tasks. The WSI thumbnails and corresponding questions from the test set were submitted to the proprietary LVLMs, which were asked to select the answer they deemed correct. The comparison results are presented in \cref{suptab:closed-cls}. As shown, the accuracy of the proprietary LVLMs did not exceed 70\% on any task, with many tasks yielding accuracies below 50\%. In contrast, OmniPath achieved over 70\% accuracy across all tasks, with most exceeding 85\%. These results indicate that OmniPath outperforms current SOTA proprietary models in pathology diagnosis tasks.

These SOTA LVLMs possess a certain degree of object detection capability, enabling them to output bounding box coordinates through language-based responses. This functionality is explicitly documented and illustrated with examples in Gemini's official documentation. For ChatGPT 4o, its object detection ability has also been confirmed through interactive user queries in its web application. We submitted the images and corresponding questions from the test set to each model via API, prompting them to perform object detection based on the question. Additionally, we appended an extra prompt to enforce a specific output format and normalized all coordinates to integers within the range of 0 to 1000. We selected tasks involving the detection of lymph nodes (LD), vessels (VD), and nerves (ND). Compared to nucleus detection, these targets are larger in size and fewer in number, making the tasks relatively easier. However, as shown in \cref{suptab:closed-det}, the performance of ChatGPT 4o and Gemini 2.5 Pro on these tasks remains nearly unusable. In contrast, OmniPath demonstrates a level of performance on these tasks that is already of practical utility.

Additionally, to validate the claim regarding OmniPath's reasoning ability illustrated in \cref{fig:sample}, we invited pathologists to annotate cancer cell nuclei located within blood vessels in several pathology images. We then evaluated ChatGPT 4o, Gemini 2.5 Pro, and OmniPath on the task of detecting intravascular cancer cell nuclei. Successfully completing this task requires the model not only to recognize vessels and distinguish between normal and cancerous nuclei, but also to reason about the spatial positions and relationships among various relevant structures in the image. The evaluation results are shown in the last column of \cref{suptab:closed-det}. As observed, OmniPath achieved the best performance on this task, while the other two models yielded results that were nearly unusable. This provides supporting evidence for the claim made in \cref{fig:sample}.

\begin{table}[h]
\centering
\caption{Accuracy (\%) comparison on part of the slide-level pathology diagnostic tasks with SOTA closed-source large vision language models.}
\label{suptab:closed-cls}
\resizebox{0.75\textwidth}{!}{
\begin{tabular}{llllll}
\toprule
Model \textbackslash{} Task                 & HCC-S & HCC-G & ICC-S & ICC-G & RCC-S \\
\midrule
chatgpt-4o-latest            & 40.23 & 37.53 & 61.22 & 35.56 & 57.58 \\
gemini-2.5-pro-preview-05-06 & 40.97 & 29.17 & 69.39 & 51.61 & 51.52 \\
OmniPath                     & \textbf{98.40} & \textbf{70.83} & \textbf{87.76} & \textbf{99.08} & \textbf{87.88} \\
\bottomrule
\end{tabular}
}
\end{table}

\begin{table}[h]
\centering
\caption{Detection performance comparison on part of the patch-level detection tasks with SOTA closed-source large vision language models.}
\label{suptab:closed-det}
\resizebox{0.9\textwidth}{!}{
\begin{tabular}{llcccc}
\toprule
Model                                         & Metric        & LD             & VD             & ND             & Cancer Nuclei in Vessels \\ \midrule
\multirow{2}{*}{chatgpt-4o-latest}            & F1-Score (\%) & 17.58          & 2.92           & 1.79           & 1.68                     \\
                                              & IoU (\%)      & 65.81          & 62.45          & 60.23          & 58.92                    \\ \midrule
\multirow{2}{*}{gemini-2.5-pro-preview-05-06} & F1-Score (\%) & 27.03          & 5.12           & 8.76           & 1.72                     \\
                                              & IoU (\%)      & 64.91          & 62.85          & 60.04          & 58.67                    \\ \midrule
\multirow{2}{*}{OmniPath}                     & F1-Score (\%) & \textbf{82.70} & \textbf{89.52} & \textbf{72.50} & \textbf{46.05}           \\
                                              & IoU (\%)      & \textbf{87.34} & \textbf{83.00} & \textbf{79.75} & \textbf{77.97}           \\
\bottomrule
\end{tabular}
}
\end{table}

\section{Dataset Sources and Construction Methods}
\label{supsec:dataset}

In this section, we will introduce the sources of various data used for training OmniPath, as well as the construction methods for dialogue data. OmniPath addresses various pathology-related tasks by receiving human queries and providing responses. Consequently, a task is typically completed in the form of a single-turn or multi-turn dialogue.

First of all, we summarize the number of samples for each organ type in the dataset in the \cref{suptab:dataset-organ} below. As the dataset contains some public samples with indeterminate organ origin, we categorize these under the "unsure" class.

\begin{table}[h]
\centering
\caption{Number of samples per organ category, with indeterminate cases labeled as ``unsure''.}
\label{suptab:dataset-organ}
\resizebox{\textwidth}{!}{
\centering
\begin{tabular}{cccccccc}
\toprule
Liver     & Stomach    & Lung          & Breast   & Kidney    & Skin   & Colon                & Brain    \\
104831    & 73259      & 71748         & 25694    & 23416     & 20251  & 13803                & 13040    \\ \midrule
Esophagus & Lymph Node & Thyroid       & Prostate & Uterus    & Ovary  & Head-Neck            & Pancreas \\
5359      & 4725       & 4315          & 3329     & 2857      & 2423   & 2306                 & 2290     \\ \midrule
Bladder   & Testis     & Adrenal Gland & Cervix   & Bile Duct & Unsure & \multicolumn{1}{l}{} & TOTAL    \\
1621      & 1563       & 1476          & 1334     & 820       & 108806 & \multicolumn{1}{l}{} & 489266   \\ 
\bottomrule
\end{tabular}
}
\end{table}

For cancer subtyping and grading tasks, we collected pathology slide WSIs of various cancers from multiple institutions. For these slides, the model is first tasked with identifying the organ of origin, using prompts provided in \cref{suptab:organ}. During dialogue data generation, a prompt is randomly selected from \cref{suptab:organ} as the human query, and the model responds with the corresponding organ label of the WSI. Following this, the model determines the disease type or pathological type using prompts listed in \cref{suptab:disease}, with two response formats: open-ended and closed-ended. In the open-ended format, the model directly provides the corresponding type, while in the closed-ended format, options are appended to the query, and the model selects the correct answer from the options. In addition, seven-class gastric lesion recognition (GLR-7) uses the same prompts to identify disease types. This task involves a large collection of gastric lesion slides gathered from multiple institutions. For cancer subtyping and grading tasks, prompts from \cref{suptab:subtyping} and \cref{suptab:grading} are used, also employing both open-ended and closed-ended response formats.

For cancer region detection and segmentation tasks, we used cancer regions annotated by physicians on the aforementioned pathology slide WSIs for training. The prompts for cancer region detection and segmentation are provided in \cref{suptab:cancer_detection} and \cref{suptab:cancer_segmentation}, respectively. The response format for the detection task is set in an XML-like format as ``\texttt{<bbox\_list><bbox>x1, y1, x2, y2</bbox>...</bbox\_list>}'', where \texttt{(x1, y1)} and \texttt{(x2, y2)} represent the relative coordinates of the top-left and bottom-right corners of a bounding box, ranging from 0 to 1 with three decimal places. Each \texttt{<bbox>} represents a bounding box, and \texttt{<bbox\_list>} stores all detection results.
For the segmentation task, the physicians' annotated cancer region boundaries are converted into polygons, with the textual response format defined as ``\texttt{<contour\_list><polygon>[x1, y1], [x2, y2], ...</polygon>...</contour\_list>}'', where each \texttt{<polygon>} contains the coordinates of vertices for one region boundary, and \texttt{<contour\_list>} stores all segmented regions. When using a mask decoder to generate segmentation results, this format is converted into corresponding masks for model predictions. This approach ensures that the generated dialogue data can accommodate models both with and without mask decoder modules.

For detection and segmentation tasks involving structures such as blood vessels, nerves, and lymph nodes, we collected pathology images containing these structures at different magnifications from multiple institutions. Physicians annotated these structures using bounding boxes or masks. The prompts used for blood vessel and lymph node detection are provided in \cref{suptab:vd} and \cref{suptab:ld}, respectively, with response formats similar to those for cancer region detection. The prompts for nerve detection and segmentation are listed in \cref{suptab:nd} and \cref{suptab:ns}, respectively, and their response formats are analogous to those used for cancer region detection and segmentation.

For the tasks of identifying microvascular invasion, neural invasion, and lymph node metastasis, we collected healthy blood vessels, nerves, and lymph nodes as negative samples, and blood vessels, nerves, and lymph nodes containing cancer cells as positive samples. The prompts used for these three tasks are listed in \cref{suptab:mvi}, \cref{suptab:ni}, and \cref{suptab:lnm}, respectively. Positive and negative samples are labeled with ``yes'' and ``no'' responses, respectively.
Additionally, for microvascular invasion, we required the model to detect cancer cell nuclei within blood vessels, using the prompt in \cref{suptab:mvid}. For neural invasion and lymph node metastasis, the model was further tasked with segmenting the corresponding cancerous regions, with prompts provided in \cref{suptab:nis} and \cref{suptab:lnms}. The response formats for these detection and segmentation tasks are consistent with those used for cancer region detection and segmentation.

For nucleus detection, we collected several publicly available nucleus segmentation datasets and converted their segmentation masks into corresponding detection bounding boxes. The segmentation datasets without class annotations include MoNuSeg~\cite{MoNuSeg}, CoNIC~\cite{conic}, and TNBC\_dataset~\cite{tnbc}, while those with class annotations include NuCLS~\cite{NuCLS} and PanNuke~\cite{PanNuke}. Prompts for datasets without class annotations are listed in \cref{suptab:ncnd}, with response formats identical to those for cancer region detection.
For datasets with class annotations, the prompts are provided in \cref{suptab:cnd}. The response format is defined as:
\begin{verbatim}
<detection_result>
  <bbox_list class="CLASS_NAME">
    <bbox>x1, y1, x2, y2</bbox>
    ...
  </bbox_list>
  ...
</detection_result>
\end{verbatim}
where the \texttt{class} attribute of \texttt{<bbox\_list>} differentiates between classes.

For other patch-level tasks, we constructed dialogues using corresponding public datasets. For organ classification, all images with a clear organ of origin were used for training, and the NuInsSeg~\cite{NuInsSeg} dataset was used for testing, with the query prompts listed in \cref{suptab:organ}. Tissue classification was conducted using the public ESCA~\cite{ESCA} dataset, with prompts corresponding to \cref{suptab:tissue}, employing both open-ended and closed-ended formats. Tumor-infiltrating lymphocyte recognition and microsatellite instability identification were performed using public datasets \cite{TIL} and \cite{MSI}, respectively, with prompts phrased as yes-no questions.
For the 32-class pan-cancer classification task, we used the TCGA-Uniform-Tumor~\cite{pancancer} dataset. Due to the large number of images in this dataset, we performed stratified sampling to extract a subset for training and testing. The prompts used correspond to \cref{suptab:disease}.
For slide-level prognosis prediction tasks in liver cancer and colorectal cancer, we collected corresponding prognosis follow-up data from multiple hospitals. The data were categorized into two classes: recurrence within two years and no recurrence within five years. A closed-ended question format was used, where the model was asked to determine which prognosis category the outcome belonged to.

In addition to the tasks directly related to pathological clinical diagnosis mentioned above, we integrated training sets from datasets such as PathInstruct~\cite{pathasst}, Quilt-Instruct-107k~\cite{quiltllava}, and PathVQA~\cite{pathvqa} to supplement and strengthen OmniPath's pathology visual question answering and instruction-following capabilities. Together, these datasets form a training dataset comprising 21 organs and approximately 490,000 training samples, significantly enhancing the model's ability to extract multi-scale and fine-grained pathological features.

% ######################### Tables of Prompts #########################
\begin{table}[h]
\centering
\caption{The list of prompts for organ identification.}
\label{suptab:organ}
\begin{tabular}{@{}p{0.98\textwidth}@{}} % 设置表格列宽度
\hline
\vspace{-0.6em} % 减少顶部间距
\begin{itemize}[label=\textbullet, left=0em, itemsep=0em, topsep=0em]
    \item What kind of organ does this image show?
    \item Classify this organ sample
    \item What organ is this?
    \item Determine organ shown
    \item Identify this organ
    \item Name the organ in the image
    \item Could you specify the organ in the picture?
    \item What kind of organ is visible in this pathology image?
    \item Identify the organ presented in this histological image.
    \item Determine the organ category in this histopathological slide.
    \item Indicate the organ observed in this pathology slide.
    \item Classify the organ depicted in this pathological slide.
    \item Examine the image and identify the organ in this histological section.
    \item Discern the organ shown in this pathology photograph.
\end{itemize} \\[-1em] % 在这里用 [-1em] 缩减换行后的垂直间距
\hline
\end{tabular}
\end{table}

\begin{table}[h]
\centering
\caption{The list of prompts for disease or pathological type classification.}
\label{suptab:disease}
\begin{tabular}{@{}p{0.98\textwidth}@{}} % 设置表格列宽度
\hline
\vspace{-0.6em} % 减少顶部间距
\begin{itemize}[label=\textbullet, left=0em, itemsep=0em, topsep=0em]
    \item Diagnose the disease from this image.
    \item Analyze this image to determine the patient's disease.
    \item Use this image to diagnose the patient's illness.
    \item What disease could this pathology slide be from?
    \item What is the pathological type?
    \item Identify the pathological type.
    \item What type of pathology is shown?
    \item Can you determine the pathology type in this image?
    \item What is the specific pathological type in this picture?
    \item Please identify the pathological type depicted in the image.
    \item Can you classify the pathological type visible in this slide?
    \item Based on the image, what is the pathology type?
    \item Could you analyze the image and determine the pathology type?
    \item Please provide a detailed analysis and identify the pathological type shown in this image.
\end{itemize} \\[-1em] % 在这里用 [-1em] 缩减换行后的垂直间距
\hline
\end{tabular}
\end{table}

\begin{table}[h]
\centering
\caption{The list of prompts for cancer subtyping.}
\label{suptab:subtyping}
\begin{tabular}{@{}p{0.98\textwidth}@{}} % 设置表格列宽度
\hline
\vspace{-0.6em} % 减少顶部间距
\begin{itemize}[label=\textbullet, left=0em, itemsep=0em, topsep=0em]
    \item Identify the cancer subtype.
    \item What is the cancer subtype?
    \item Can you determine the cancer subtype?
    \item What cancer subtype is shown in this image?
    \item Please identify the cancer subtype in this image.
    \item Can you classify the cancer subtype visible in this slide?
    \item What is the specific cancer subtype depicted in this picture?
    \item Could you determine the cancer subtype based on this image?
    \item Analyze the image and identify the cancer subtype.
    \item Please provide a detailed analysis and identify the cancer subtype shown in this pathology image.
    \item Identify the histological subtype.
    \item What is the histological subtype?
    \item Can you determine the histological subtype?
    \item What histological subtype is shown in this image?
    \item Please identify the histological subtype in this image.
    \item Can you classify the histological subtype visible in this slide?
    \item What is the specific histological subtype depicted in this picture?
    \item Could you determine the histological subtype based on this image?
    \item Analyze the image and identify the histological subtype.
    \item Please provide a detailed analysis and identify the histological subtype shown in this pathology image.
\end{itemize} \\[-1em] % 在这里用 [-1em] 缩减换行后的垂直间距
\hline
\end{tabular}
\end{table}

\begin{table}[h]
\centering
\caption{The list of prompts for cancer grading.}
\label{suptab:grading}
\begin{tabular}{@{}p{0.98\textwidth}@{}} % 设置表格列宽度
\hline
\vspace{-0.6em} % 减少顶部间距
\begin{itemize}[label=\textbullet, left=0em, itemsep=0em, topsep=0em]
    \item Grade the cancer in this image.
    \item What is the grade of cancer shown in this picture?
    \item Can you determine the cancer grade in this image?
    \item Identify the grade of cancer visible in this image.
    \item Please analyze and grade the cancer depicted in this image.
    \item Could you assess and indicate the grade of cancer in this picture?
    \item Examine this image and provide the cancer grade.
    \item Can you evaluate and classify the cancer severity shown in this image?
    \item Please examine the cancer cells in this image and determine their differentiation grade.
    \item Carefully analyze the differentiation of the cancer cells in this image and provide a detailed grading based on their appearance.
    \item Identify the histological grade.
    \item What is the histological grade?
    \item Can you determine the histological grade?
    \item What histological grade is shown in this image?
    \item Please identify the histological grade in this image.
    \item Can you classify the histological grade visible in this slide?
    \item What is the specific histological grade depicted in this picture?
    \item Could you determine the histological grade based on this image?
    \item Analyze the image and identify the histological grade.
    \item Please provide a detailed analysis and identify the histological grade shown in this pathology image.
\end{itemize} \\[-1em] % 在这里用 [-1em] 缩减换行后的垂直间距
\hline
\end{tabular}
\end{table}

\begin{table}[h]
\centering
\caption{The list of prompts for cancer region detection.}
\label{suptab:cancer_detection}
\begin{tabular}{@{}p{0.98\textwidth}@{}} % 设置表格列宽度
\hline
\vspace{-0.6em} % 减少顶部间距
\begin{itemize}[label=\textbullet, left=0em, itemsep=0em, topsep=0em]
    \item Does this image have any cancer areas? If so, provide the bounding boxes for each.
    \item Are there cancer regions in this picture? Please give bounding boxes for any cancer areas.
    \item Can you identify cancer in this image? If present, list the bounding boxes of the cancer areas.
    \item Check this image for cancer areas and give me the bounding boxes if there are any.
    \item Is cancer visible in this image? If yes, outline the cancer areas with bounding boxes.
    \item Answer yes or no: Does this pathology image have cancer? If yes, provide bounding boxes for the cancer areas.
    \item Is there cancer in this pathology image? If so, give me the bounding boxes for the cancerous regions.
    \item Can you detect cancer in this pathology image? Yes or no, and if yes, indicate the cancer areas with bounding boxes.
    \item Please confirm whether this pathology image contains cancer. Provide bounding boxes for any cancer areas.
    \item Does this pathology image show any cancer regions? If it does, outline these areas with bounding boxes.
    \item Does this pathology image contain cancer? If so, provide bounding boxes for each area in [x1, y1, x2, y2] format with coordinates normalized between 0 and 1, up to three decimal places.
    \item Is there cancer in this pathology picture? If yes, list the cancer regions' bounding boxes as [x1, y1, x2, y2], with normalized coordinates and three decimal accuracy.
    \item Can you identify cancer areas in this pathology image? Please give their bounding boxes in the format [x1, y1, x2, y2], with normalized 0 to 1 coordinates, precise to three decimals.
    \item Check for cancer in this pathology image and provide the bounding boxes of any found, in the format [x1, y1, x2, y2], with coordinates normalized from 0 to 1 and rounded to three decimal places.
    \item Are there any cancerous regions in this pathology image? If present, outline them using bounding boxes in the format [x1, y1, x2, y2], with normalized coordinates (0 to 1 scale) and three decimal point precision.
\end{itemize} \\[-1em] % 在这里用 [-1em] 缩减换行后的垂直间距
\hline
\end{tabular}
\end{table}

\begin{table}[h]
\centering
\caption{The list of prompts for cancer region segmentation.}
\label{suptab:cancer_segmentation}
\begin{tabular}{@{}p{0.98\textwidth}@{}} % 设置表格列宽度
\hline
\vspace{-0.6em} % 减少顶部间距
\begin{itemize}[label=\textbullet, left=0em, itemsep=0em, topsep=0em]
    \item Does this image have any cancer areas? If so, provide the bounding boxes for each.
    \item Are there cancer regions in this picture? Please give bounding boxes for any cancer areas.
    \item Can you identify cancer in this image? If present, list the bounding boxes of the cancer areas.
    \item Check this image for cancer areas and give me the bounding boxes if there are any.
    \item Is cancer visible in this image? If yes, outline the cancer areas with bounding boxes.
    \item Answer yes or no: Does this pathology image have cancer? If yes, provide bounding boxes for the cancer areas.
    \item Is there cancer in this pathology image? If so, give me the bounding boxes for the cancerous regions.
    \item Can you detect cancer in this pathology image? Yes or no, and if yes, indicate the cancer areas with bounding boxes.
    \item Please confirm whether this pathology image contains cancer. Provide bounding boxes for any cancer areas.
    \item Does this pathology image show any cancer regions? If it does, outline these areas with bounding boxes.
    \item Does this pathology image contain cancer? If so, provide bounding boxes for each area in [x1, y1, x2, y2] format with coordinates normalized between 0 and 1, up to three decimal places.
    \item Is there cancer in this pathology picture? If yes, list the cancer regions' bounding boxes as [x1, y1, x2, y2], with normalized coordinates and three decimal accuracy.
    \item Can you identify cancer areas in this pathology image? Please give their bounding boxes in the format [x1, y1, x2, y2], with normalized 0 to 1 coordinates, precise to three decimals.
    \item Check for cancer in this pathology image and provide the bounding boxes of any found, in the format [x1, y1, x2, y2], with coordinates normalized from 0 to 1 and rounded to three decimal places.
    \item Are there any cancerous regions in this pathology image? If present, outline them using bounding boxes in the format [x1, y1, x2, y2], with normalized coordinates (0 to 1 scale) and three decimal point precision.
\end{itemize} \\[-1em] % 在这里用 [-1em] 缩减换行后的垂直间距
\hline
\end{tabular}
\end{table}

\begin{table}[h]
\centering
\caption{Prompt list for blood vessel detection.}
\label{suptab:vd}
\begin{tabular}{@{}p{0.98\textwidth}@{}} % 设置表格列宽度
\hline
\vspace{-0.6em} % 减少顶部间距
\begin{itemize}[label=\textbullet, left=0em, itemsep=0em, topsep=0em]
    \item Detect all vessels.
    \item Find every blood vessel.
    \item Identify all vessels in the image.
    \item Locate all blood vessels.
    \item Can you detect all blood vessels in this image?
    \item Could you show all the vessels in the image?
    \item Locate and mark every blood vessel in this picture.
    \item Please identify and create bounding boxes around every blood vessel visible in this image, including both large and small vessels.
\end{itemize} \\[-1em] % 在这里用 [-1em] 缩减换行后的垂直间距
\hline
\end{tabular}
\end{table}

\begin{table}[h]
\centering
\caption{Prompt list for lymph node detection.}
\label{suptab:ld}
\begin{tabular}{@{}p{0.98\textwidth}@{}} % 设置表格列宽度
\hline
\vspace{-0.6em} % 减少顶部间距
\begin{itemize}[label=\textbullet, left=0em, itemsep=0em, topsep=0em]
    \item Detect all lymph nodes.
    \item Find all lymph nodes in this image.
    \item Identify and mark all lymph nodes present in the pathology image.
    \item Can you detect and highlight every lymph node in this pathology slide?
\end{itemize} \\[-1em] % 在这里用 [-1em] 缩减换行后的垂直间距
\hline
\end{tabular}
\end{table}

\begin{table}[h]
\centering
\caption{Prompt list for nerve detection.}
\label{suptab:nd}
\begin{tabular}{@{}p{0.98\textwidth}@{}} % 设置表格列宽度
\hline
\vspace{-0.6em} % 减少顶部间距
\begin{itemize}[label=\textbullet, left=0em, itemsep=0em, topsep=0em]
    \item Detect all nerves.
    \item Find all nerves in this image.
    \item Identify and mark all nerves present in the pathology image.
    \item Can you detect all nerves in this pathology slide?
    \item Please locate and highlight every nerve visible in this pathology image.
\end{itemize} \\[-1em] % 在这里用 [-1em] 缩减换行后的垂直间距
\hline
\end{tabular}
\end{table}

\begin{table}[h]
\centering
\caption{Prompt list for nerve segmentation.}
\label{suptab:ns}
\begin{tabular}{@{}p{0.98\textwidth}@{}} % 设置表格列宽度
\hline
\vspace{-0.6em} % 减少顶部间距
\begin{itemize}[label=\textbullet, left=0em, itemsep=0em, topsep=0em]
    \item Segment all nerves.
    \item Can you segment the nerves in this image?
    \item Identify and segment all nerves present in the pathology image.
    \item Please detect and segment all nerves in this pathology slide.
    \item Could you locate, identify, and segment every nerve visible in this pathology image?
\end{itemize} \\[-1em] % 在这里用 [-1em] 缩减换行后的垂直间距
\hline
\end{tabular}
\end{table}

\begin{table}[h]
\centering
\caption{Prompt list for microvascular invasion identification.}
\label{suptab:mvi}
\begin{tabular}{@{}p{0.98\textwidth}@{}} % 设置表格列宽度
\hline
\vspace{-0.6em} % 减少顶部间距
\begin{itemize}[label=\textbullet, left=0em, itemsep=0em, topsep=0em]
    \item Is this MVI?
    \item Does this image show MVI?
    \item Can you confirm if this is an example of microvascular invasion?
    \item Based on this image, would you say this represents microvascular invasion?
    \item Considering the details in this image, could you analyze and determine whether it illustrates microvascular invasion?
\end{itemize} \\[-1em] % 在这里用 [-1em] 缩减换行后的垂直间距
\hline
\end{tabular}
\end{table}

\begin{table}[h]
\centering
\caption{Prompt list for neural invasion identification.}
\label{suptab:ni}
\begin{tabular}{@{}p{0.98\textwidth}@{}} % 设置表格列宽度
\hline
\vspace{-0.6em} % 减少顶部间距
\begin{itemize}[label=\textbullet, left=0em, itemsep=0em, topsep=0em]
    \item Is this neural invasion?
    \item Does this image show neural invasion?
    \item Can you confirm if this image represents neural invasion?
    \item Based on this image, is it indicative of neural invasion?
    \item Could you analyze this image and determine if it depicts neural invasion?
\end{itemize} \\[-1em] % 在这里用 [-1em] 缩减换行后的垂直间距
\hline
\end{tabular}
\end{table}

\begin{table}[h]
\centering
\caption{Prompt list for lymph node metastasis identification.}
\label{suptab:lnm}
\begin{tabular}{@{}p{0.98\textwidth}@{}} % 设置表格列宽度
\hline
\vspace{-0.6em} % 减少顶部间距
\begin{itemize}[label=\textbullet, left=0em, itemsep=0em, topsep=0em]
    \item Is this lymph node metastasis?
    \item Does this image show lymph node metastasis?
    \item Can you confirm if this image represents lymph node metastasis?
    \item Based on this image, is it indicative of lymph node metastasis?
    \item Could you analyze this image and determine if it depicts lymph node metastasis?
\end{itemize} \\[-1em] % 在这里用 [-1em] 缩减换行后的垂直间距
\hline
\end{tabular}
\end{table}

\begin{table}[h]
\centering
\caption{Prompt list for detection of cancer cell nuclei within vessels.}
\label{suptab:mvid}
\begin{tabular}{@{}p{0.98\textwidth}@{}} % 设置表格列宽度
\hline
\vspace{-0.6em} % 减少顶部间距
\begin{itemize}[label=\textbullet, left=0em, itemsep=0em, topsep=0em]
    \item Please identify all cancerous nuclei in this vessel.
    \item Detect every cancerous cell nucleus present in the vessel.
    \item Identify all the cancerous nuclei within this blood vessel.
    \item Find and mark all cancerous cell nuclei in the vessel.
    \item Locate every cancerous nucleus in this blood vessel.
    \item Detect all cancerous nuclei in the vessel using bounding boxes.
    \item Identify every cancerous nucleus in the vessel and mark them with bbox.
    \item Please use bbox to outline all cancerous cell nuclei present in this vessel.
    \item Find all the cancerous nuclei in the vessel and use bounding boxes for each.
    \item Locate and mark every cancerous cell nucleus in the blood vessel with a bbox.
\end{itemize} \\[-1em] % 在这里用 [-1em] 缩减换行后的垂直间距
\hline
\end{tabular}
\end{table}

\begin{table}[h]
\centering
\caption{Prompt list for segmentation of cancerous regions in neural invasion.}
\label{suptab:nis}
\begin{tabular}{@{}p{0.98\textwidth}@{}} % 设置表格列宽度
\hline
\vspace{-0.6em} % 减少顶部间距
\begin{itemize}[label=\textbullet, left=0em, itemsep=0em, topsep=0em]
    \item Segment the cancerous area in the nerve.
    \item Can you segment the cancerous region in this nerve?
    \item Please identify and segment the cancerous areas within this nerve.
    \item Could you analyze and segment all the cancerous regions in the nerve shown in this image?
    \item Can you detect and segment the specific areas of cancer within the nerve in this pathology image?
\end{itemize} \\[-1em] % 在这里用 [-1em] 缩减换行后的垂直间距
\hline
\end{tabular}
\end{table}

\begin{table}[h]
\centering
\caption{Prompt list for segmentation of cancerous regions in lymph nodes.}
\label{suptab:lnms}
\begin{tabular}{@{}p{0.98\textwidth}@{}} % 设置表格列宽度
\hline
\vspace{-0.6em} % 减少顶部间距
\begin{itemize}[label=\textbullet, left=0em, itemsep=0em, topsep=0em]
    \item Segment the cancerous area in the lymph node.
    \item Can you segment the cancerous region in this lymph node?
    \item Please identify and segment the cancerous areas within this lymph node.
    \item Could you analyze and segment all the cancerous regions in the lymph node shown in this image?
\end{itemize} \\[-1em] % 在这里用 [-1em] 缩减换行后的垂直间距
\hline
\end{tabular}
\end{table}

\begin{table}[h]
\centering
\caption{Prompt list for nucleus detection without class label.}
\label{suptab:ncnd}
\begin{tabular}{@{}p{0.98\textwidth}@{}} % 设置表格列宽度
\hline
\vspace{-0.6em} % 减少顶部间距
\begin{itemize}[label=\textbullet, left=0em, itemsep=0em, topsep=0em]
    \item Please identify all nuclei in this image.
    \item Detect every cell nucleus present in the picture.
    \item Identify all the nuclei within this image.
    \item Find and mark all nuclei in the image.
    \item Locate every nucleus in this picture.
    \item Detect all cell nuclei in the image using bounding boxes.
    \item Identify every nucleus in the picture and mark them with bbox.
    \item Please use bbox to outline all nuclei present in this image.
    \item Find all the cell nuclei in the image and use bounding boxes for each.
    \item Locate and mark every nucleus in the picture with a bbox.
    \item Detect all nuclei in this pathology image and output with bounding boxes in [x1, y1, x2, y2] format, normalized coordinates to 0-1, accurate to three decimals.
    \item Identify every cell nucleus in the picture, marking them with bbox in [x1, y1, x2, y2], normalize coordinates between 0 and 1, with three decimal precision.
    \item Please use bbox to indicate all nuclei in this image, with coordinates in [x1, y1, x2, y2] format, normalized to 0-1 and rounded to three decimal places.
    \item Find all nuclei in the pathology image and represent each with a bounding box, using [x1, y1, x2, y2] for normalized coordinates to a scale of 0 to 1, with three digits after the decimal.
    \item Locate every nucleus in this image, using bbox for output in [x1, y1, x2, y2] format, with coordinates normalized to 0-1, and precision up to three decimals.
\end{itemize} \\[-1em] % 在这里用 [-1em] 缩减换行后的垂直间距
\hline
\end{tabular}
\end{table}

\begin{table}[h]
\centering
\caption{Prompt list for nucleus detection with class label.}
\label{suptab:cnd}
\begin{tabular}{@{}p{0.98\textwidth}@{}} % 设置表格列宽度
\hline
\vspace{-0.6em} % 减少顶部间距
\begin{itemize}[label=\textbullet, left=0em, itemsep=0em, topsep=0em]
    \item Please detect and classify all nuclei in this image.
    \item Detect and classify every cell nucleus present in the picture.
    \item Detect and classify all the nuclei within this image.
    \item Detect and classify all nuclei in the image.
    \item Locate every nucleus and give its category in this picture.
    \item Detect all cell nuclei in the image using bounding boxes with labels.
    \item Detect and classify every nucleus in the picture and mark them with bbox.
    \item Please use bbox to outline all nuclei and indicate every label present in this image.
    \item Distinguish all the cell nuclei in the image and use bounding boxes for each.
    \item Detect and classify every nucleus in the picture with a bbox.
    \item Detect and classify all nuclei in this pathology image and output with bounding boxes in [x1, y1, x2, y2] format, normalized coordinates to 0-1, accurate to three decimals.
    \item Identify every cell nucleus with label in the picture, marking them with bbox in [x1, y1, x2, y2], normalize coordinates between 0 and 1, with three decimal precision.
    \item Please use bbox to detect and classify all nuclei in this image, with coordinates in [x1, y1, x2, y2] format, normalized to 0-1 and rounded to three decimal places.
    \item Find all nuclei in the pathology image and represent each with a bounding box and a category, using [x1, y1, x2, y2] for normalized coordinates to a scale of 0 to 1, with three digits after the decimal.
    \item Locate and classify every nucleus in this image, using bbox for output in [x1, y1, x2, y2] format, with coordinates normalized to 0-1, and precision up to three decimals.
\end{itemize} \\[-1em] % 在这里用 [-1em] 缩减换行后的垂直间距
\hline
\end{tabular}
\end{table}

\begin{table}[h]
\centering
\caption{Prompt list for tissue identification.}
\label{suptab:tissue}
\begin{tabular}{@{}p{0.98\textwidth}@{}} % 设置表格列宽度
\hline
\vspace{-0.6em} % 减少顶部间距
\begin{itemize}[label=\textbullet, left=0em, itemsep=0em, topsep=0em]
    \item Identify the tissue type in this image.
    \item What is the tissue type shown in this picture?
    \item Can you determine the tissue type in this image?
    \item Identify the type of tissue visible in this image.
    \item Please analyze and identify the tissue type depicted in this image.
    \item Could you assess and indicate the tissue type in this picture?
    \item Examine this image and provide the tissue type.
    \item Can you evaluate and classify the tissue type shown in this image?
    \item Please examine the tissue cells in this image and determine their type.
    \item Carefully analyze the tissue cells in this image and provide a detailed identification based on their appearance.
\end{itemize} \\[-1em] % 在这里用 [-1em] 缩减换行后的垂直间距
\hline
\end{tabular}
\end{table}
% ####################### End Tables of Prompts #######################

% ####################### Appendix Reference #######################
\clearpage
% {
%     \small
%     \bibliography{appendix}
%     \bibliographystyle{ieeenat_fullname}
% }
\printbibliography[title={Appendix References}, heading=bibintoc] % 这里不能是input，否则附录的Reference会和正文一样，且实现附录和正文reference需要chapterbib包

%%%%%%%%%%%%%%%%%%%%%%%%%%%%%%%%%%%%%%%%%%%%%%%%%%%%%%%%%%%%%%%%%%%%%%

\end{document}